\documentclass[10pt,journal,compsoc]{IEEEtran}
\ifCLASSOPTIONcompsoc
  \usepackage[nocompress]{cite}
\else
  \usepackage{cite}
\fi
\ifCLASSINFOpdf
   \usepackage[pdftex]{graphicx}
\else
   \usepackage[dvips]{graphicx}
\fi
\usepackage{array}
\usepackage{fixltx2e}

\usepackage{stfloats}
\usepackage{url}

\usepackage{bm}
\usepackage{subfigure}
\usepackage{amsmath,amssymb,amsthm}
\usepackage{multirow,array}
\usepackage{color}
\usepackage{algorithmic,algorithm}
\usepackage{lipsum}
\usepackage{booktabs}
\usepackage{graphicx}

\usepackage{diagbox}
\usepackage{arydshln}
\usepackage{bbding}
\usepackage[colorlinks,linkcolor=blue]{hyperref}

\begin{document}
%
\title{LibFewShot: A Comprehensive Library for Few-shot Learning}
%
%
%
%

\author{Wenbin~Li,
        Ziyi Wang,
        Xuesong Yang,
        Chuanqi Dong,
        Pinzhuo Tian,
        Tiexin Qin,
        Jing Huo,
        Yinghuan Shi,
        Lei~Wang,~\IEEEmembership{Senior Member,~IEEE,}
        Yang~Gao,
        and~Jiebo~Luo,~\IEEEmembership{Fellow,~IEEE}
\IEEEcompsocitemizethanks{\IEEEcompsocthanksitem Wenbin Li, Ziyi Wang, Xuesong Yang, Chuanqi Dong, Jing Huo, Yinghuan Shi and Yang Gao are with the State Key Laboratory for Novel Software Technology, Nanjing University, China, 210023 (e-mail: \{liwenbin, huojing, syh, gaoy\}@nju.edu.cn; \{wangziyi, yangxuesong, dongchuanqi\}@smail.nju.edu.cn).
\IEEEcompsocthanksitem Pinzhuo Tian is with the School of Computer Engineering and Science, Shanghai University, China (e-mail: pinzhuo@shu.edu.cn).
\IEEEcompsocthanksitem Tiexin Qin is with the Department of Electrical Engineering, City University of Hong Kong, China (e-mail: tiexinqin@gmail.com).
\IEEEcompsocthanksitem Lei Wang is with the School of Computing and Information Technology, University of Wollongong, Australia (e-mail: leiw@uow.edu.au).
\IEEEcompsocthanksitem Jiebo Luo is with the Department of Computer Science, University of Rochester, America (e-mail: jluo@cs.rochester.edu).
    }}
\IEEEtitleabstractindextext{%
\begin{abstract}
Few-shot learning, especially few-shot image classification, has received increasing attention and witnessed significant advances in recent years. Some recent studies implicitly show that many generic techniques or ``tricks'', such as data augmentation, pre-training, knowledge distillation, and self-supervision, may greatly boost the performance of a few-shot learning method. Moreover, different works may employ different software platforms, backbone architectures and input image sizes, making fair comparisons difficult and practitioners struggle with reproducibility. To address these situations, we propose a comprehensive library for few-shot learning (LibFewShot) by re-implementing eighteen state-of-the-art few-shot learning methods in a unified framework with the same single codebase in PyTorch. Furthermore, based on LibFewShot, we provide comprehensive evaluations on multiple benchmarks with various backbone architectures to evaluate common pitfalls and effects of different training tricks. In addition, with respect to the recent doubts on the necessity of meta- or episodic-training mechanism, our evaluation results confirm that such a mechanism is still necessary especially when combined with pre-training. We hope our work can not only lower the barriers for beginners to enter the area of few-shot learning but also elucidate the effects of nontrivial tricks to facilitate intrinsic research on few-shot learning. The source code is available from 
\href{https://github.com/RL-VIG/LibFewShot}{https://github.com/RL-VIG/LibFewShot}.
\end{abstract}

}

\maketitle

\IEEEdisplaynontitleabstractindextext

%
\IEEEpeerreviewmaketitle

\IEEEraisesectionheading{\section{Introduction}\label{sec:introduction}}

%
%
%
%

\IEEEPARstart{F}ew-shot learning (FSL), especially few-shot image classification, has received considerable attention in recent years~\cite{fei2006one,vinyals2016matching,finn2017model,sung2018learning,sun2020meta,simon2020dsn,LiWHSGL20_ADM,ye2020heterogeneous,yang2021free,abbas2022sharp}. It tries to learn an effective classification model from a few labeled training examples. A wide variety of advanced FSL methods has been proposed and significantly improved the classification performance on multiple benchmark datasets~\cite{gidaris2018dynamic,lee2019meta,sun2019meta,li2019distribution,li2019revisiting,ye2020few,xie2022joint,afrasiyabi2022matching}.

Because of the extreme scarcity of training examples per class, it is almost impossible to only use the few available examples to learn an effective classifier to solve the few-shot classification problem. Therefore, the current FSL methods generally follow a paradigm of transfer learning, \emph{i.e.,} using a large labeled but class-disjoint auxiliary set to learn transferable knowledge (or representations) to boost the target few-shot task. More importantly, different from standard transfer learning~\cite{pan2009survey}, the existing FSL methods normally adopt a meta-training~\cite{santoro2016meta} or episodic-training mechanism~\cite{vinyals2016matching} to train a few-shot model by constructing massive few-shot tasks (episodes) from the auxiliary set to simulate the target few-shot task. Like the target few-shot task, each simulated task also consists of a labeled support (training) set and an unlabeled query (test) set.

In general, these typical FSL methods can be roughly divided into two types, \emph{i.e.,} meta-learning based~\cite{santoro2016meta,ravi2017optimization} and metric-learning based~\cite{koch2015siamese,snell2017prototypical}. The former normally adopts a meta-learning or learning-to-learn paradigm~\cite{thrun1998lifelong,vilalta2002perspective} to learn some kind of cross-task knowledge through an alternate optimization between the meta-learner and base-learner. In this way, it is able to make the model quickly generalize to new unseen tasks with a few training examples (\emph{i.e.,} test time fine-tuning). In contrast, the latter employs a learning-to-compare paradigm~\cite{vinyals2016matching} to learn representations that can be transferred between tasks without test time fine-tuning (test-tuning for short). This is implemented by directly comparing the relations between query images and support images in each training task through an episodic training mechanism~\cite{vinyals2016matching}. Both types of methods have greatly advanced the development of few-shot learning.

However, the question ``\textit{Is meta- or episodic-training paradigm really crucial and optimal for the FSL problem?}'' has been raised recently by the community. Some recent works~\cite{BertinettoHTV19_R2D2,ChenLKWH19,dhillon2020baseline,chen2020new,TianWKTI20_RFS} have attempted to answer this question. For example, \cite{ChenLKWH19} finds that a simple baseline method, \emph{i.e.,} pre-training $+$ test-tuning (called non-episodic based methods), can obtain competitive results when compared with the meta- or episodic-training paradigm based methods. Similarly, \cite{TianWKTI20_RFS} proposes an improved baseline method by using logistic regression as the linear classifier in the test-tuning phase, which surprisingly achieves the state of the art. Therefore, \textit{can we conclude that the meta- or episodic-training paradigm is indeed NOT necessary for FSL?}


In addition, we observe that the implementation and evaluation details of different FSL methods vary significantly. This could make fair comparison difficult, or even worse, make some conclusions questionable. Specifically, certain key discrepancies of existing FSL methods can be summarized as follows: (1) different software platforms (\textit{e.g.,} TensorFlow vs. PyTorch); (2) different backbones (\textit{e.g.,} Wider ResNet12 vs. ResNet12); (3) classifiers with different parameters (\textit{e.g.,} heavy-parametric classifier vs. non-parametric classifier); (4) different input image sizes (\textit{e.g.,} $224\times224$ vs. $84\times84$); (5) different test time evaluations (\textit{e.g.,} center crop evaluation vs. raw evaluation). Clearly, such wide discrepancies are not amenable to a fair comparison and therefore cannot truthfully reflect the actual progress of FSL. Moreover, some FSL methods may employ additional fancy deep learning tricks, such as stronger data augmentation, knowledge distillation, self-supervision, label smoothing, and DropBlock, in the training process. The effects of different deep learning tricks are worth studying in FSL. Also, the additional used tricks are another key type of discrepancy.

\textbf{Our work.} Therefore, to facilitate fair comparison and conveniently investigate the common issues in FSL, we develop \textit{a comprehensive library for few-shot learning (LibFewShot)} by making most of the implementation details of different FSL methods consistent. To be specific, eighteen representative state-of-the-art FSL methods, including seven meta-learning based, six metric-learning based and five non-episodic based methods, are systematically re-implemented in a unified framework with the same single codebase in PyTorch. In LibFewShot, we try our best to ensure that all the methods use the same settings and the same bag of tricks, except for some specific tricks or specific neural architectures that are the main contributions of certain methods. In this way, we could take a true picture of the actual state-of-the-art results on FSL. More importantly, we are able to construct a large-scale study to analyze the impact of different tricks, such as pre-training, global classification, knowledge distillation and label smoothing, in a fair way.

\textbf{Our contributions.} The main contributions of this work are as follows:
\begin{itemize}
    \item We develop a unified framework LibFewShot with eighteen re-implemented FSL methods for the first time in the literature. It can be used as a toolbox and a platform to help practitioners efficiently use and reproduce FSL methods. 
    \item We provide comprehensive evaluations of the eighteen methods on multiple benchmark datasets with various embedding backbones, by controlling the implementation details. This can reveal the actual progress of FSL and can be conveniently referenced to perform comparative experiments.
    \item We conduct a large-scale study on multiple representative FSL methods in a fair way, revealing that (1) pre-training indeed can learn a good initial representation but is not necessarily an optimal representation; (2) meta- or episodic-training can further improve this initial representation; (3) $\ell_2$ normalization of image feature vectors can significantly boost the final classification more than test-tuning in the test phase, especially in data-limited scenarios.
    \item We conduct comprehensive ablation studies for multiple deep learning tricks on the FSL problem, showing that many tricks could achieve significant algorithm-agnostic performance improvements and are universally applicable to different FSL methods.
    \item We have released LibFewShot as an open-source project on GitHub, and will continue to add new methods into this project. We welcome other researchers to contribute to this library to facilitate the community to conduct research on this important topic together.
\end{itemize}

\begin{figure*}[!tp]
\centering
\subfigure[Non-episodic based methods]{
           \includegraphics[height=0.175\textheight]{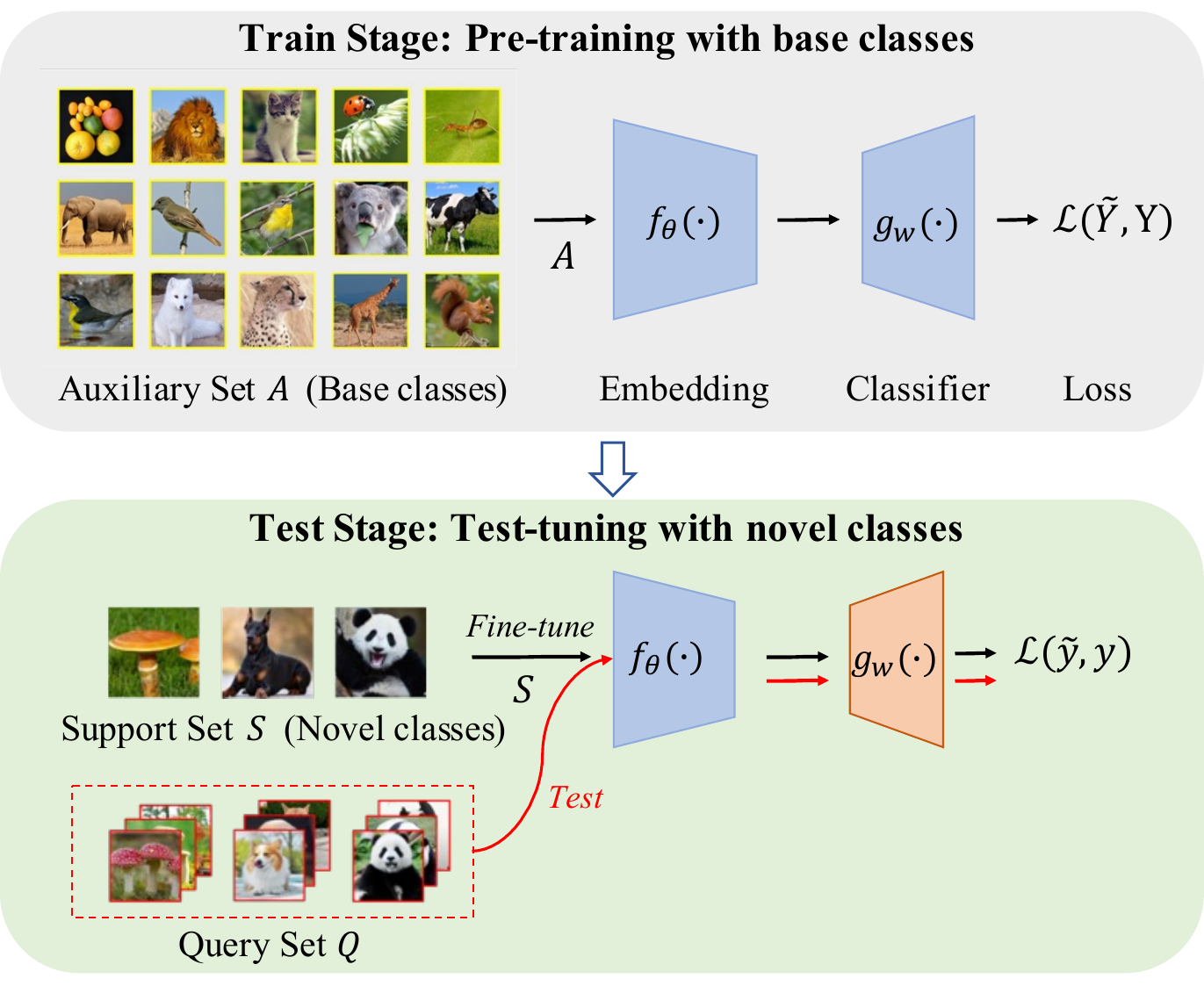}}
\subfigure[Meta-learning based methods]{
           \includegraphics[height=0.175\textheight]{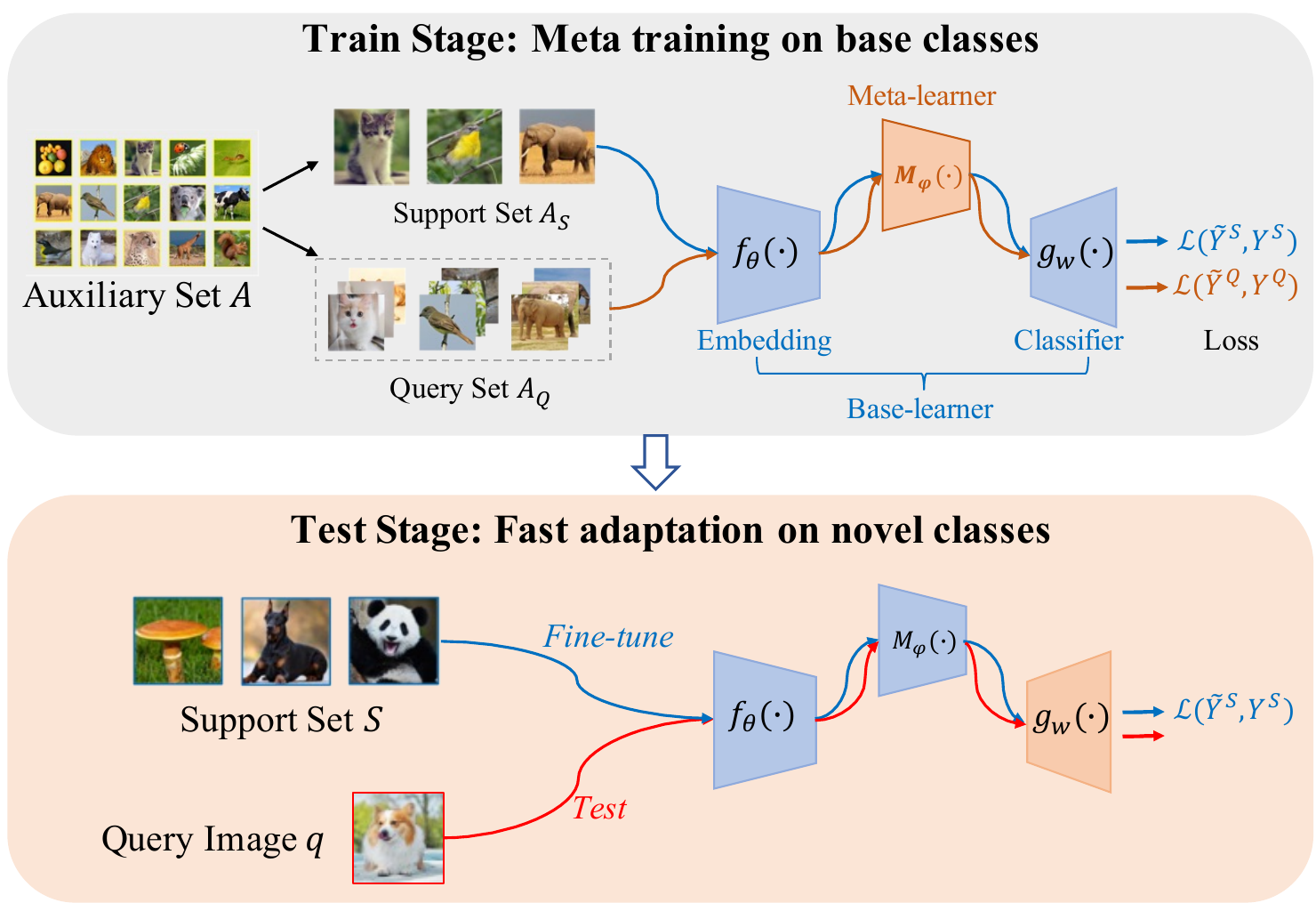}}
\subfigure[Metric-learning based methods]{
           \includegraphics[height=0.175\textheight]{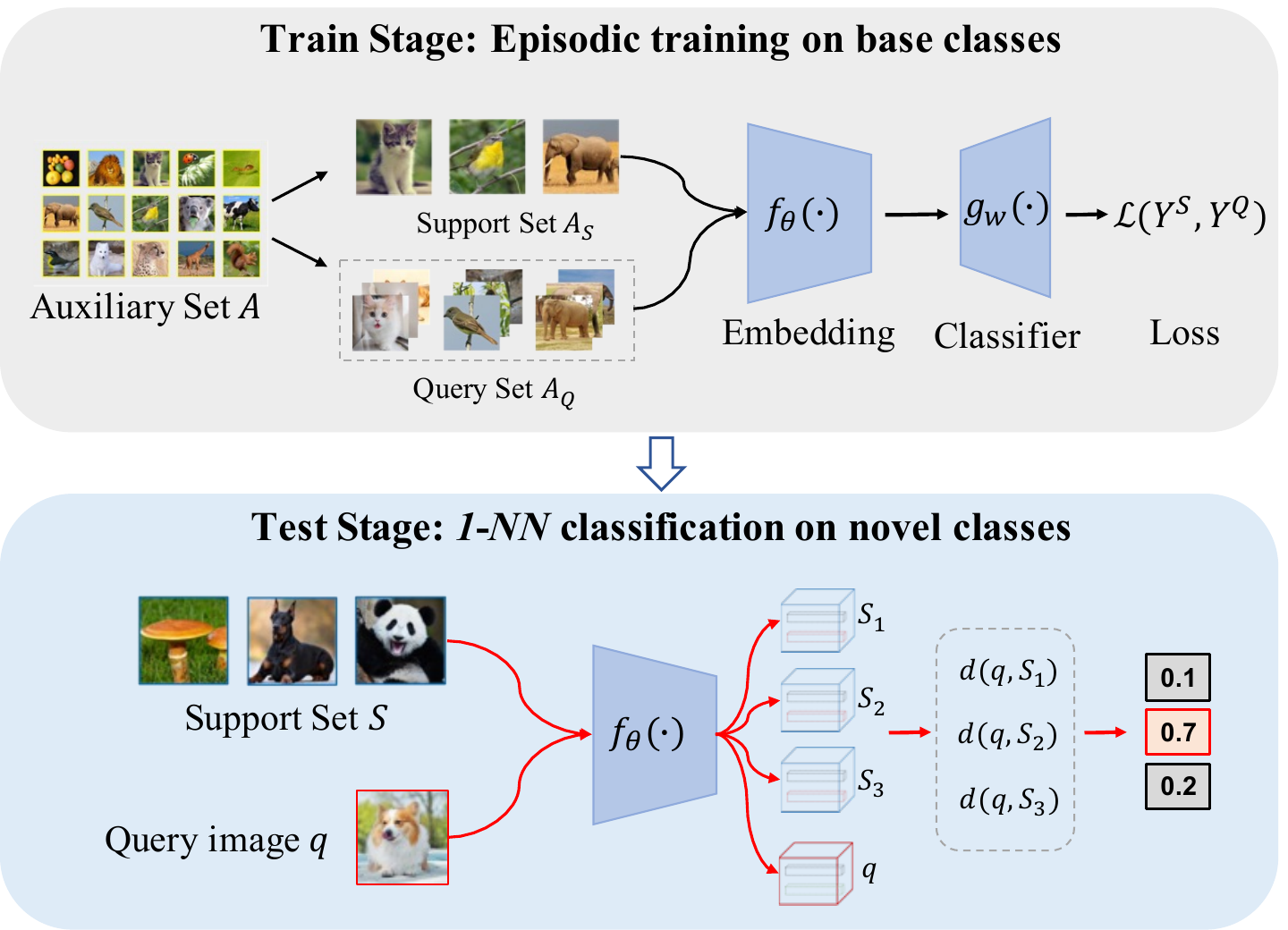}}
\caption{Illustrations of \textit{Non-episodic based}, \textit{Meta-learning based} and \textit{Metric-learning based methods}, respectively. The first one uses a generic classification task as the proxy task, while the latter two use a meta- or episodic-training paradigm in the training stage. At the test stage, the first two will test-tune a new task-relevant classifier, while the last one simply uses a $1$-NN classifier without any test-tuning.}
\label{fig_flowchart}
\end{figure*}

\section{Overview of Few-shot Learning Methods}
\label{Section_3}
In this section, we will first introduce the problem formulation of FSL, and then review three kinds of FSL methods, \textit{i.e.,} non-episodic based methods, meta-learning based methods and metric-learning based methods, where multiple representative methods are further reviewed in detail.

\subsection{Problem Formulation}
In few-shot setting, there are usually three sets of data, including a target labeled support set $\mathcal{S}$, a target unlabeled query set $\mathcal{Q}$ and a class-disjoint auxiliary set $\mathcal{A}$. In particular, $\mathcal{S}$ and $\mathcal{Q}$ share the same label space, which corresponds to the training and test sets in generic classification, respectively. The concept of ``\textit{few-shot}'' in fact comes from $\mathcal{S}$, where there are $\mathcal{C}$ classes but each class only has $\mathcal{K}$ (\emph{e.g.,} $1$ or $5$) labeled samples. We call this kind of classification task a $\mathcal{C}$-way $\mathcal{K}$-shot task. Clearly, such a few labeled samples in each class make it almost impossible to train an effective classification model, no matter using deep neural networks or traditional machine learning algorithms. Therefore, one solution of FSL becomes how to use $\mathcal{A}$ to boost the learning on the target task (\emph{i.e.,} $\mathcal{S}$ and $\mathcal{Q}$). The good point is that $\mathcal{A}$ generally enjoys more classes and samples per class than $\mathcal{S}$, while the challenge is that $\mathcal{A}$ has a disjoint label space from $\mathcal{S}$ and even may have a large domain shift from $\mathcal{S}$.

Therefore, the current FSL methods mainly focus on how to effectively learn transferable knowledge from $\mathcal{A}$ for fast adaptation (\emph{e.g.,} meta-learning based and non-episodic based methods) or for good generalization (\emph{e.g.,} metric-learning based methods) on $\mathcal{S}$ with a few labeled support examples. Note that, in experiment at study, given a dataset $\mathcal{D}$, it will be divided into $\mathcal{D}_{train}$, $\mathcal{D}_{val}$ and $\mathcal{D}_{test}$ for training, validation and test, respectively. Typically, $\mathcal{D}_{train}$ will be taken as the auxiliary set $\mathcal{A}$, and multiple evaluation few-shot tasks $\mathcal{T}=\langle\mathcal{S}, \mathcal{Q}\rangle$ will be formed by randomly sampling from $\mathcal{D}_{val}$ and $\mathcal{D}_{test}$, respectively.

\textbf{Notation.} Following the literature, the auxiliary set, \emph{i.e.,} the set of base classes, is denoted as $\mathcal{A}=\{X_i, y_i\}_{i=1}^N$, with the image $X_i\in \mathbb{R}^{H\times W\times 3}$ and the one-hot labeling vector $y_i\in Y=\{0,1\}^{\mathcal{C}_{base}}$. For a $\mathcal{C}$-way $\mathcal{K}$-shot task with $\mathcal{C}$ novel classes, the support set and query set are represented as $\mathcal{S}=\{S_1,\cdots,S_\mathcal{C}\}=\{X_i, y_i\}_{i=1}^{\mathcal{CK}}$ and $\mathcal{Q}=\{Q_i, y_i\}_{i=1}^{\mathcal{CM}}$, respectively, where $S_c=\{X_i, y_c\}_{i=1}^{\mathcal{K}}$ contains $\mathcal{K}$ images and is the $c$-th class in $\mathcal{S}$. Let $f_\theta(\cdot)$ and $g_\omega(\cdot)$ denote the convolutional neural network based embedding backbone and classifier, respectively. Also, $g_\omega(\cdot)$ can be integrated with $f_\theta(\cdot)$ into a same network and trained in an end-to-end manner. For generic classification and the base learner in meta-learning, the cost function is represented as $\mathcal{L}\big(g_\omega(f_\theta(X)), y\big)$. As for the metric-learning based FSL, the cost function is represented as $\mathcal{L}\big(g_\omega(f_\theta(X) | \mathcal{S}), y\big)$.

\textbf{Episodic-training and Meta-training.}
To learn an effective FSL model, meta-training~\cite{santoro2016meta} or episodic-training mechanism~\cite{vinyals2016matching} is normally adopted at the training stage. Both meta- and episodic-training rely on a lot of simulation few-shot tasks, which are randomly constructed from the auxiliary set $\mathcal{A}$. Each simulated task $\mathcal{T}$ consists of two subsets, $\mathcal{A_{S}}$ and $\mathcal{A_{Q}}$, which are akin to $\mathcal{S}$ and $\mathcal{Q}$, respectively. Note that because the labels in each simulated task are randomly assigned according to the original real labels in the auxiliary set, we call such a kind of label as \textit{local-label}. In contrast, if the real labels of the auxiliary set are directly used, we mean that \textit{global-label} is used.

At each iteration, one simulated task (episode), \emph{i.e.,} $\mathcal{T}=\langle\mathcal{A}_\mathcal{S}, \mathcal{A}_\mathcal{Q}\rangle$, is adopted to train the current model. Conceptually, tens of thousands of tasks, \emph{i.e.,} $\{\mathcal{T}^i=\langle\mathcal{A}^i_\mathcal{S}, \mathcal{A}^i_\mathcal{Q}\rangle\}_{i=1}^\mathcal{M}\in \rho(\mathcal{T})$, will be randomly sampled from a task distribution $\rho(\mathcal{T})$ to train this model. The core principle is that the training condition (\emph{i.e.,} the training task) must match the test condition (\emph{i.e.,} the target task)~\cite{vinyals2016matching}. From the perspective of meta-learning, as more tasks are observed, the model can use the accumulated meta-knowledge to adapt its own bias according to the characteristics of each task~\cite{thrun1998learning,vilalta2002perspective}.

\subsection{Non-episodic based Methods}
As illustrated in Figure~\ref{fig_flowchart}(a), the \textit{non-episodic based methods}~\cite{ChenLKWH19,TianWKTI20_RFS,rajasegaran2020self_SKD,dhillon2020baseline,Mangla20_S2M2_mixup,yang2021free} generally follow the standard transfer learning procedure~\cite{pan2009survey}, consisting of two phases, \emph{i.e.,} pre-training with base classes and test-tuning with novel classes.

\textbf{Pre-training with base classes.} In this phase, the whole auxiliary set $\mathcal{A}$ is used to train a $\mathcal{C}_{base}$-class classifier by using the standard cross-entropy loss as below,
\begin{equation}\label{fun1}
\begin{split}
 \mathnormal{\Gamma}=\underset{\mathcal{\theta,\omega}}{\arg\min}\sum_{i=1}^N\mathcal{L}^\text{CE}\Big(g_\omega(f_\theta(X_i)),y\Big)\,,
\end{split}
\end{equation}
where $\mathcal{L}^\text{CE}$ is the cross-entropy loss function.

\textbf{Test-tuning with novel classes.} Test-tuning is performed in the test phase. Specifically, for each specific novel task $\mathcal{T}=\langle\mathcal{S}, \mathcal{Q}\rangle$, a new $\mathcal{C}$-class classifier will be re-learned based on $\mathcal{S}$ every time. Basically, the pre-trained embedding parameter $\theta$ is fixed to avoid over-fitting, because there is limited labeled data in $\mathcal{S}$. Once the novel classifier is learned, the labels of $\mathcal{Q}$ can be predicted.

\textbf{Representative methods} include \textit{Baseline}~\cite{ChenLKWH19}, \textit{Baseline++}~\cite{ChenLKWH19}, RFS-simple~\cite{ChenLKWH19}, SKD-GEN0~\cite{rajasegaran2020self_SKD}, S2M2~\cite{Mangla20_S2M2_mixup} and Neg-Cosine~\cite{Negative-cosine_ECCV20}, \emph{etc.} The main difference between the first three methods is that they use different classifiers at the test-tuning stage: (1) Baseline~\cite{ChenLKWH19} adopts a linear layer, \emph{i.e.,} a fully-connected (FC) layer, as the new classifier; (2) Baseline++~\cite{ChenLKWH19} replaces the standard inner product (in the FC layer) with a cosine distance between the input feature and weight vector; (3) RFS-simple~\cite{ChenLKWH19} employs logistic regression instead of the FC layer as the new classifier by first using $\ell_2$ normalization for the feature vector.

SKD-GEN0~\cite{rajasegaran2020self_SKD} also uses logistic regression as the classifier as RFS-simple~\cite{ChenLKWH19}, where the only difference is that additional rotation-based self-supervision~\cite{rotation_2018} is further introduced into the pre-training stage. In addition, both~\cite{ChenLKWH19} and~\cite{rajasegaran2020self_SKD} develop an extended version, respectively, by using knowledge distillation~\cite{hinton2015_KD}. As for S2M2, more auxiliary tasks, such as Manifold Mixup~\cite{manifoldMixup2019}, rotation~\cite{rotation_2018} and Exemplar~\cite{Exemplar-2014}, are introduced into the pre-training stage to learn more powerful representations. Neg-Cosine~\cite{Negative-cosine_ECCV20} introduces a negative margin loss, \emph{i.e.,} a negative-margin cosine softmax loss, at the pre-training stage, and shows that this will benefit the novel classes in the test-tuning phase.

\textbf{Discussions.} The above non-episodic based methods have achieved surprisingly good results with a much simpler methodology, shaking the foundation of the current pure meta- or episodic-training based methods. The question of ``\textit{Should we discard meta- or episodic-training in FSL?}" will be interesting to investigate. On the other hand, intuitively, the cross-entropy loss used in the pre-training stage may make the learned representations overfit the seen base classes, thus lacking generalization ability for unseen classes. Moreover, the non-episodic methods strictly follow the paradigm of standard transfer learning and heavily focus on improving the pre-training stage by utilizing the latest and popular deep learning tricks, which may somewhat overlook the intrinsic problems of FSL.

\subsection{Meta-learning based Methods}
As illustrated in Figure~\ref{fig_flowchart}(b), \textit{meta-learning based methods}~\cite{finn2017model,gordon2018versa,lee2019meta,BertinettoHTV19_R2D2,RusuRSVPOH19_LEO,raghu2020ANIL,xu2020attentional} normally perform a meta-training paradigm on a family of few-shot tasks constructed from the base classes at the training stage, aiming to make the learned model able to quickly adapt to unseen novel tasks at the test stage. In particular, the meta-training procedure consists of a two-step optimization between the base-learner and meta-learner. 
Specifically, given a sampled task $\mathcal{T}=\langle\mathcal{A}_\mathcal{S}, \mathcal{A}_\mathcal{Q}\rangle$, Step-1 (\emph{i.e.,} base-learning or inner loop) is to use $\mathcal{A}_\mathcal{S}$ (\emph{i.e.,} training examples in each task) to learn the base-learner. Next, in Step-2 (\emph{i.e.,} meta-tuning or outer loop), $\mathcal{A}_\mathcal{Q}$ (\emph{i.e.,} test samples in each task) is employed to optimize the meta-learner. In this way, the meta-learner is expected to learn a kind of across-task meta-knowledge, which can be used for the fast adaptation on novel tasks.

\textbf{Model-Agnostic Meta-Learning (MAML)} is one representative method~\cite{finn2017model}, whose core idea is to train a model's initial parameters by involving the second-order gradients, making this model able to rapidly adapt to a new task just with one or a few gradient steps. Specifically, in the base-learning phase (inner loop), given $\mathcal{T}=\langle\mathcal{A}_\mathcal{S}, \mathcal{A}_\mathcal{Q}\rangle$, the current model $F_\varTheta=f_\theta\circ g_\omega$ and $\varTheta=\varTheta_0$, we can obtain the $m$-th inner loop gradient update as,
\begin{equation}\label{fun2}
\begin{split}
 \varTheta_m=\varTheta_{m-1} -\alpha\nabla_{\varTheta_{\scriptsize{m-1}}}\mathcal{L}_{\mathcal{A}_\mathcal{S}}(F_{\varTheta_{m-1}})\,,
\end{split}
\end{equation}
where $\varTheta=\{\theta, \omega\}$, $\alpha$ is a step size hyper-parameter, and $m$ is the total number of inner iterations. Next, in the meta-tuning phase (outer loop), the parameter of the model is truly updated over the previous parameter $\varTheta$ rather than $\varTheta_m$ by using the query set $\mathcal{A}_\mathcal{Q}$, \emph{i.e.,}
\begin{equation}\label{fun3}
\begin{split}
 \varTheta=\varTheta -\beta\nabla_\varTheta\mathcal{L}_{\mathcal{A}_\mathcal{Q}}(F_{\varTheta_m})\,,
\end{split}
\end{equation}
where $\beta$ is a meta step size hyper-parameter.

\textbf{Ridge Regression Differentiable Discriminator (R2D2)} is designed from another perspective~\cite{BertinettoHTV19_R2D2}, by adopting a standard machine learning algorithm such as ridge regression as the base-learner classifier $g_\omega(\cdot)$ in the inner loop. Note that the base-learner classifier in MAML is a standard FC layer. The advantage of R2D2 is that ridge regression enjoys a closed-form solution, which can make the base-learning phase more efficient. Specifically, the ridge regression with parameter matrix $\bm{W}\in\mathbb{R}^{d\times c}$ is formulated as,
\begin{equation}\label{fun4}
\begin{split}
 \mathnormal{\Gamma}&=\underset{\bm{W}}{\arg\min}\|\bm{X}\bm{W}-\bm{Y}\|^2+\lambda\|\bm{W}\|^2 \\
                    &=(\bm{X}^\top\bm{X}+\lambda \bm{I})^{-1}\bm{X}^\top\bm{Y}\,,
\end{split}
\end{equation}
where $\bm{X}\in\mathbb{R}^{n\times d}$ and $\bm{Y}\in\mathbb{R}^{n\times c}$ denote $n$ input samples with $d$-dimensional features and the corresponding labels (\emph{i.e.,} $c$ classes), respectively, $\bm{I}$ is the identity matrix, and $\lambda$ is a regularization hyper-parameter.

Specifically, suppose there are $n$ support images in $\mathcal{A}_\mathcal{S}$, and $f_\theta(\cdot)$ can be used to obtain the feature embeddings, \emph{i.e.,} $\bm{X}=f_\theta(\mathcal{A}_\mathcal{S})\in\mathbb{R}^{n\times d}$. In the base-learning phase, the optimal parameter matrix $\bm{W}^\ast$ can be easily obtained according to Eq.(\ref{fun4}). Next, in the meta-tuning phase, the predictions of $\bm{X}_\mathcal{Q}=f_\theta(\mathcal{A}_\mathcal{Q})\in\mathbb{R}^{n\times d}$ can be achieved as,
\begin{equation}\label{fun5}
\begin{split}
 \widehat{\bm{Y}}=\alpha\bm{X}_\mathcal{Q}\bm{W}^\ast+\beta\,,
\end{split}
\end{equation}
where $\alpha$ and $\beta$ are scale and bias, respectively, which can be learned by optimizing the meta-loss $\mathcal{L}(\widehat{\bm{Y}},\bm{Y}_\mathcal{Q})$.

\textbf{Other Representative Methods} include \textit{Latent Embedding Optimization (LEO)}~\cite{RusuRSVPOH19_LEO}, \textit{Almost No Inner Loop (ANIL)}~\cite{raghu2020ANIL}, \textit{Body Only update in Inner Loop (BOIL)}~\cite{OhYKY2021BOIL}, \textit{MetaOptNet}~\cite{lee2019meta} and \textit{Versa}~\cite{gordon2018versa}. Specifically, LEO, ANIL and BOIL all follow the same optimization procedure as MAML. The core idea of LEO is to optimize the meta-learning process within a low-dimensional latent space, and to learn a generative distribution of model parameters, instead of directly learning the explicit high-dimensional model parameters in MAML. ANIL tries to remove the inner loop updates for the embedding backbone (\emph{i.e.,} $f_\theta$), but only applies the inner loop adaptation to the classifier (\emph{i.e.,} $g_\omega$), which means that only $\omega$ is updated in Eq.(\ref{fun2}). In contrast with ANIL, BOIL updates only the embedding backbone $f_\theta$ but freezes the update of the classifier $g_\omega$ in the inner loop. Similar to R2D2, MetaOptNet also attempts to use a convex base learner, \emph{i.e,} linear support vector machine (SVM), in the inner loop for FSL. Different from the above four methods, Versa is designed from a new perspective of Bayesian learning by introducing a versatile amortization network.

\textbf{Discussions.} We can see that there are mainly two development directions in the meta-learning based methods: (1) an implicit two-loop optimization direction following the procedure of MAML, such as LEO, ANIL and BOIL; (2) an explicit two-loop optimization direction following the procedure of R2D2, \emph{e.g.} MetaOpeNet. The former follows a trend of designing a more efficient optimization-based meta-learning method, aiming to address the complicated optimization problem of meta-training. The latter aims to design a more effective base learner by introducing the traditional and classic machine learning algorithms into the paradigm of meta learning. On the other hand, note that the early meta-learning based methods mainly employ the pure meta-training paradigm to learn a model from scratch. Some methods, such as MTL~\cite{sun2019meta} and LEO~\cite{RusuRSVPOH19_LEO}, have already introduced pre-training into the training process, because the pre-training technique can be easily leveraged as pre-processing. Therefore, the effect of pre-training is worth further investigating in meta-learning based FSL methods.

\subsection{Metric-learning based Methods}
As illustrated in Figure~\ref{fig_flowchart}(c), different from the two-loop structure of meta-learning based methods, \textit{metric-learning based methods}~\cite{koch2015siamese,snell2017prototypical,sung2018learning,li2019revisiting,DoerschGZ20,zhang2020deepemd,wertheimer2021few,kang2021relational} directly compare the similarities (or distances) between the query images and support classes (\emph{i.e.,} learning-to-compare) through \textit{one single feed-forward pass} through the episodic-training mechanism~\cite{vinyals2016matching}. In other words, for each input query image, the entire support set $\mathcal{A}_\mathcal{S}$ is jointly encoded into the latent embedding space simultaneously, and their relationships (\emph{i.e.,} outputs) are used to classify. In this way, \emph{i.e.,} by conditioning on the support set, it is able to enable the model adapt to the characteristics of each task, and make the learned representations transferable between different tasks.

\textbf{Prototypical Networks (ProtoNet)} is a typical metric-learning based method~\cite{snell2017prototypical}, which takes the mean vector of each support class as its corresponding prototype representation, and then compares the relationships between the query image and prototypes. Specifically, given a few-shot task $\mathcal{T}=\langle\mathcal{A}_\mathcal{S}, \mathcal{A}_\mathcal{Q}\rangle$, $\mathcal{A}_\mathcal{S}=\{S_1, S_2, \cdots, S_C\}$, the prototype $\bm{c}_i\in\mathbb{R}^d$ of each class $S_i$ can be formulated as,
\begin{equation}\label{fun6}
\begin{split}
 \bm{c}_i= \frac{1}{|S_i|}\sum_{j=1}^\mathcal{K} f_\theta(X_j)\,,
\end{split}
\end{equation}
where $X_j\in S_i$, $|S_i|=\mathcal{K}$ denotes there are $\mathcal{K}$ images (\emph{i.e.,} $\mathcal{K}$-shot) in the $i$-th support class. Here, $f_\theta(X_j)\in\mathbb{R}^d$ means $f_\theta(\cdot)$ extracts $d$-dimensional global feature representation for each input image. Given a distance function $D(\cdot,\cdot)$, such as Euclidean distance, the predicted posterior probability distributions of a query image $Q$ is,
\begin{equation}\label{fun7}
\begin{split}
 \rho(y=i|Q)= \frac{\exp\big(-D(f_\theta(Q),\bm{c}_i)\big)}{\sum_{j=1}^C\exp\big(-D(f_\theta(Q),\bm{c}_j)\big)}\,.
\end{split}
\end{equation}
Specifically, at the training stage, the standard cross-entropy loss can be employed to train the entire model. Also, during test, the nearest-neighbor classifier ($1$-NN) can be conveniently used for prediction.

\textbf{Deep Nearest Neighbor Neural Network (DN4)} is another representative method~\cite{li2019revisiting}, which argues that performing pooling on local features into a compact global-level representation will lose considerable discriminative information. Instead, DN4 advocates to directly use the raw local features and employs a local descriptor based \textit{image-to-class (I2C)} measure to learn transferable local features.
Specifically, given an input image $X$, without the last pooling or FC layer of the embedding network, $f_\theta(X)\in\mathbb{R}^{d\times h\times w}$ will be a three-dimensional tensor, and can be reshaped as a set of $d$-dimensional local descriptors
\begin{equation}\label{fun8}
    f_\theta(X)=[\bm{x}_1,\ldots,\bm{x}_n]\in \mathbb{R}^{d \times n}\,,
\end{equation}
where $\bm{x}_i$ is the $i$-th local descriptor and $n=h\times w$ is the total number of local descriptors for image $X$. Note that, in ProtoNet, both query image and the images in the support classes are represented with a global feature vector, respectively. Especially, the class prototype is also an average-pooling of multiple global feature vectors (\emph{e.g.,} the $\mathcal{K}$-shot setting). In contrast, in DN4, both query image and each support class are represented with a set of local descriptors without any pooling.

Suppose a query image $Q$ and a support class $S$ are represented as $f_\theta(Q)=[\bm{x}_1,\ldots,\bm{x}_n]\in \mathbb{R}^{d \times n}$ and $f_\theta(S)=[f_\theta(X_1),\ldots,f_\theta(X_K)]\in \mathbb{R}^{d \times nK}$, respectively. The image-to-class measure will be calculated as 
\begin{equation}\label{function_5}\small
\begin{split}
    D_\text{I2C}(Q,S) & = \sum_{i=1}^{n}\text{Topk}\Big(\frac{f_\theta(Q)^\top\cdot f_\theta(S)}{\|f_\theta(Q)\|_F\cdot\|f_\theta(S)\|_F}\Big)\,,
\end{split}
\end{equation}
where $\|\cdot\|_F$ denotes the Frobenius norm, and $\text{Topk}(\cdot)$ means selecting the $k$ largest elements in each row of the correlation matrix between $Q$ and $S$.

\begin{table*}[!tp]\footnotesize
	\centering
	\caption{\textbf{Data splits used in each dataset}. $\mathcal{C}_\text{all}$/$\mathcal{N}_\text{train}$ is the total number of classes/images. $\mathcal{C}_\text{train}$/$\mathcal{N}_\text{train}$, $\mathcal{C}_\text{val}$/$\mathcal{N}_\text{val}$ and $\mathcal{C}_\text{test}$/$\mathcal{N}_\text{test}$ indicate the number of classes/images in training (auxiliary), validation and test sets, respectively.}
	\begin{tabular}{lccccc}
	\toprule
	\textbf{Dataset} & \textbf{\textit{mini}ImageNet} & \textbf{\textit{tiered}ImageNet} &  \textbf{Stanford Dogs} & \textbf{Stanford Cars} & \textbf{CUB Birds-200-2011}  \\ 
	\midrule
	$\mathcal{C}_\text{all}$   &  $100$    &  $608$   &  $120$    &  $196$    &  $200$     \\
	$\mathcal{C}_\text{train}$ &  $64$     &  $351$   &  $70$     &  $130$    &  $130$     \\
	$\mathcal{C}_\text{val}$   &  $16$     &  $97$    &  $20$     &  $17$     &  $20$  \\
	$\mathcal{C}_\text{test}$  &  $20$     &  $160$   &  $30$     &  $49$     &  $50$     \\
	\midrule
	$\mathcal{N}_\text{all}$   &  $60000$  &  $779165$    &  $20580$  &  $16185$  &  $11788$  \\
	$\mathcal{N}_\text{train}$ &  $38400$  &  $448695$    &  $12165$  &  $10766$  &  $7648$ \\
	$\mathcal{N}_\text{val}$   &  $9600$   &  $124261$    &  $3312$   &  $1394$   &  $1182$  \\
	$\mathcal{N}_\text{test}$  &  $12000$  &  $206209$    &  $5103$   &  $4025$   &  $2958$  \\
	\bottomrule
	\end{tabular}
	\label{libfewshot_datasets}
\end{table*}

\textbf{Other Representative Methods} include \textit{Relation Network (RelationNet)}~\cite{sung2018learning}, \textit{Covariance Metric Network (CovaMNet)}~\cite{li2019distribution}, \textit{Cross Attention Network (CAN)}~\cite{hou2019cross}, \textit{Deep Earth Mover's Distance (DeepEMD)}~\cite{zhang2020deepemd}, \textit{Few-shot Embedding Adaptation with Transformer (FEAT)}~\cite{ye2020few}, and \textit{Relational Embedding Network (RENet)}~\cite{kang2021renet}, \emph{etc.} The core of RelationNet is to learn a non-linear metric through a deep convolutional neural network, instead of choosing a specific metric function, \emph{e.g.,} Euclidean distance. Instead of using traditional first-order class representations, \emph{e.g.,} mean vector, CovaMNet proposes a second-order local covariance representation to represent each class along with a new covariance metric. From the perspective of attention, CAN proposes to calculate the cross attention between each pair of class feature and query feature so as to learn more discriminative features. Similar to DN4, DeepEMD also uses the set of local descriptors as the representation for an image and employs the Earth Mover's Distance to calculate a structural distance between dense local representations of two images. FEAT proposes to take a set-to-set transformation via a transformer layer to make the global instance embedding of support set become task-specific for better adaptation. Similar to CAN, RENet proposes a self-correlational representation module and a cross-correlational attention module to learn relational patterns within and between images, respectively.

\textbf{Discussions.} From the recent advances, we can see that the main trends in metric-learning based methods are in two folds: (1) how to effectively represent each image and each support class; and (2) how to design a more powerful metric function. Specifically, the research trends show that using local descriptor representations may be a good choice and designing a task-adaptive metric function is important. In addition, we notice that because there are generally no data-dependent parameters in the classifier (\emph{i.e.,} $1$-NN classifier), the metric-learning based methods do not have the test-tuning procedure at the test stage. Therefore, is the paradigm employed by metric-learning based methods reasonable? In other words, is test-tuning really essential for the FSL problem?

\begin{table*}[!tp]\footnotesize
\extrarowheight=3pt
\tabcolsep=1pt
\centering
\caption{\textbf{Reproduction results} on \textit{mini}ImageNet using the original paper settings. Results are reported with the mean accuracy over $3000$ $5$-way $1$-shot and $5$-way $5$-shot test tasks, respectively. Global-label indicates that the global labels of the auxiliary set are used for pre-training or additional global classification during training. Local-label means that only specific local labels are used in the episodic- or meta-training phase. Test-Tune means test-tuning of using the support set during test. Test-DA denote data augmentation during test. KD means knowledge distillation and SS means self-supervision. $^\dagger$ and $^\ddagger$ indicate that the standard backbones are different or slightly modified. \textcolor{blue}{\checkmark} with blue color indicates a given trick is not used in the original setting but used in our reproduction experiment. - means the result is not reported in the original paper.}
\begin{tabular}{l l c c c  c c c c c c c c c c }
\toprule[1pt]
\multirow{2}{*}{\textbf{Method}}  &\multirow{2}{*}{\textbf{Embed.}} &\multicolumn{2}{c}{\textbf{Image Size}}    &\multicolumn{7}{c}{\textbf{Training Tricks}}                &\multicolumn{2}{c}{\textbf{$5$-way $1$-shot}}  &\multicolumn{2}{c}{\textbf{$5$-way $5$-shot}}  \\ \cmidrule{3-15}
                                                        &                    &84  &224                     &Lr/Optimizer/Decay    &Global-label  &Local-label  &Test-Tune               &Test-DA     &KD &SS     &Reported  &Ours    &Reported  &Ours\\ 
\midrule
\multirow{2}{*}{\textbf{Baseline}}       & Conv64F   &  \checkmark  &              & $0.001$/Adam/No      &  \checkmark &  &  \checkmark    &              &              &              &$42.11$     &$42.34$  &$62.53$    &$62.18$\\
                                         & ResNet18  &              &  \checkmark  & $0.001$/Adam/No      &  \checkmark &  &  \checkmark    &              &              &              &$51.75$   &$51.18$   &$74.27$    &$74.05$ \\
\cmidrule{2-15}
\multirow{2}{*}{\textbf{Baseline++}}     & Conv64F   &  \checkmark  &              & $0.001$/Adam/No      &  \checkmark & &  \checkmark    &              &              &               &$48.24$     &$46.21$   &$66.43$    &$65.18$ \\
                                         & ResNet18  &              &  \checkmark  & $0.001$/Adam/No      &  \checkmark & &  \checkmark    &              &              &               &$51.87$     &$53.60$   &$75.68$    &$73.63$ \\
\cmidrule{2-15}
\textbf{RFS-simple}                      & ResNet12  &  \checkmark  &              & $0.05$/SGD/Step       &  \checkmark & &  \checkmark    &  \checkmark  &              &               &$62.02$     & $62.80$    &$79.64$  & $79.57$   \\
\cmidrule{2-15}
\textbf{RFS-distill}                     & ResNet12  &  \checkmark  &              & $0.05$/SGD/Step       &  \checkmark & &  \checkmark    &  \checkmark  &  \checkmark  &               &$64.82$ &$63.44$ &$82.14$ &$80.17$\\
\cmidrule{2-15}
\textbf{SKD-GEN0}                        & ResNet12  &  \checkmark  &              & $0.05$/SGD/Step        &  \checkmark & &  \checkmark    &  \checkmark  &              &  \checkmark   &$65.93$ &$66.45$  & $83.15$ & $83.43$ \\
\cmidrule{2-15}
\textbf{SKD-GEN1}                        & ResNet12  &  \checkmark  &              & $0.05$/SGD/Step        &  \checkmark & &  \checkmark    &  \checkmark  &  \checkmark  &  \checkmark   &$67.04$ &$67.09$      & $83.54$ & $83.67 $ \\
\cmidrule{2-15}
\textbf{Neg-Cosine}                      & ResNet12  &  \checkmark  &              &$0.003$/Adam/Cosine       &  \checkmark & & \checkmark    &    & &                                       &$63.85$ &$63.28$      & $81.57$ & $81.24$ \\
\midrule\midrule
\multirow{1}{*}{\textbf{MAML}}           & Conv32F   &  \checkmark  &              & $0.001$/Adam/Step      &             &\checkmark  &  \checkmark       &      &      &                 & $48.70$   & $47.41$   & $63.11$  & $65.24$   \\
\cmidrule{2-15}
\textbf{Versa}                           & Conv64F$^\dagger$  & \checkmark &       & $1e^{-4}$/Adam/Step     &             &\checkmark  &  \checkmark      &           &      &             & $53.40$   & $51.92$   & $67.37$  & $66.26$   \\
\cmidrule{2-15}
\multirow{2}{*}{\textbf{R2D2}}           & Conv64F  & \checkmark   &               & $0.001$/Adam/Step     &             &\checkmark  &  \checkmark    &              &          &        & $49.50$   &$47.57$    & $65.40$  & $66.68$    \\
                                         & Conv64F$^\ddagger$ & \checkmark &       & $0.001$/Adam/Step    &             &\checkmark  &  \checkmark    &              &      &            & $51.80$   &$55.53$    & $68.40$  & $70.79$   \\
\cmidrule{2-15}
\multirow{1}{*}{\textbf{ANIL}}           & Conv32F   & \checkmark  &               & $0.001$/Adam/Step     &             &\checkmark &  \checkmark       &   &    &                       & $46.70$   & $48.44$   & $61.50$  & $64.35$   \\
\cmidrule{2-15}
\multirow{1}{*}{\textbf{LEO}}            & WRN-28-10   & \checkmark  &             & $4e^{-4}$/Adam/No     & \checkmark            &\checkmark &  \checkmark       &   &    &                       & $61.76$   & $55.89$   & $77.59$  & $70.55$   \\
\cmidrule{2-15}
\multirow{2}{*}{\textbf{BOIL}}           & Conv64F   & \checkmark  &                & $0.001$/Adam/Step    &   &\checkmark  &  \checkmark     &              &              &             &$49.61$     &$48.00$  &$66.45$    & $64.39$\\
                                         & ResNet12  &  {\checkmark} & & $0.001$/Adam/Step    &   &\checkmark  &  \checkmark    &              &              &              &$-$   &$58.87$   &$71.30$    &$72.88$\\
\cmidrule{2-15}
\multirow{2}{*}{\textbf{MTL}} & \multirow{2}{*}{ResNet12}  &\multirow{2}{*}{\checkmark} & &$0.1$/SGD/Step &\multirow{2}{*}{\checkmark} &\multirow{2}{*}{\checkmark}  &\multirow{2}{*}{\checkmark}   &  &    &    &\multirow{2}{*}{$60.20$}    & \multirow{2}{*}{$60.20$}   &\multirow{2}{*}{$74.30$}  &\multirow{2}{*}{$75.86$}   \\
                                                             &  &  &                        & $0.001$/Adam/Step      &  &  &          &   &    &    &    &    &   &     \\
\midrule\midrule
\textbf{ProtoNet}$^\dagger$              & Conv64F &  \checkmark  &                 & $0.001$/Adam/Step        &   &\checkmark &      &  &  &                                             & $46.14$  & $46.30$   & $65.77$   & $66.24$   \\
\cmidrule{2-15}
\textbf{RelationNet}                     & Conv64F & \checkmark  &                  & $0.001$/Adam/Step        &   &\checkmark  &     &   &     &                                         & $50.44$  & $51.75$  & $65.32$ & $66.77$  \\
\cmidrule{2-15}
\textbf{CovaMNet}                       & Conv64F & \checkmark &                    & $0.001$/Adam/Step        &   &\checkmark  &     &  &   &                                            & $51.19$  & $53.36$  & $67.65$  & $68.17$   \\
\cmidrule{2-15}
\multirow{2}{*}{\textbf{DN4}}           & Conv64F & \checkmark  &                   & $0.001$/Adam/Step        &   &\checkmark &      &   &   &                                           & $51.24$   & $51.95$   & $71.02$  & $71.42$   \\
                                        & ResNet12$^\dagger$ &\checkmark &          & $0.001$/Adam/Step        &   &\checkmark  &  &   &   &                                              & $54.37$   & $57.76$   & $74.44$  & $77.57$ \\
\cmidrule{2-15}
\textbf{CAN}                            & ResNet12  & \checkmark &                  & $0.1$/SGD/Step             &\checkmark & \checkmark   &   &  & &                                      & $63.85$  & $66.62$ & $79.44$  & $78.96$ \\
\cmidrule{2-15}
\textbf{RENet}                          & ResNet12  &  \checkmark  &                & $0.1$/SGD/MultiStep       &\checkmark  &\checkmark &   &    & &                                      &$67.60$ &$66.83$      & $82.58$ & $82.13$ \\
\bottomrule
\end{tabular} 
\label{Table1}
\end{table*}

\begin{table*}[!tp]\footnotesize
\extrarowheight=2pt
\tabcolsep=4pt
\centering
\caption{\textbf{The overview picture of the state of the art} on \textit{mini}ImageNet and \textit{tiered}ImageNet by controlling the most common implementation details except some special tricks, with our \textbf{LibFewShot}. The fifth column shows the total number of trainable parameters used by each method. Global-label indicates that the global labels of the auxiliary set are used for pre-training or global classification during training. Local-label means that only the specific local labels are used in the episodic- or meta-training phase. Test-tune means test-tuning of using the support set at the test stage. Note that SKD-GEN0 uses an additional self-supervision trick that is the core of this method.}
\begin{tabular}{l r c c c c c c  |c c c c}
\toprule[1pt]
\multirow{2}{*}{\textbf{Method}} & \multirow{2}{*}{\textbf{Venue}} & \multirow{2}{*}{\textbf{Embed.}} & \multirow{2}{*}{\textbf{Type}} & \multirow{2}{*}{\textbf{Para.}} & \multicolumn{3}{c|}{\textbf{Tricks}} & \multicolumn{2}{c}{\textbf{\textit{mini}ImageNet}}  & \multicolumn{2}{c}{\textbf{\textit{tiered}ImageNet}}   \\ \cmidrule{6-12}
& & & & &Global-label &Local-label &Test-tune    & $1$-shot & $5$-shot  & $1$-shot & $5$-shot\\ 
\midrule
\textbf{Baseline}~\cite{ChenLKWH19}               & ICLR'19        & Conv64F      & Non-episodic  &  $0.22$M      &  \checkmark & &  \checkmark       &  $44.90$       &  $63.96$       &  $48.20$       &  $68.96$  \\
\textbf{Baseline++}~\cite{ChenLKWH19}             & ICML'19        & Conv64F      & Non-episodic  &  $0.22$M      &  \checkmark & &  \checkmark       &  $\bm{48.54}$  &  $65.47$       &  $49.73$       &  $70.14$  \\
\textbf{RFS-simple}~\cite{TianWKTI20_RFS}         & ECCV'20        & Conv64F      & Non-episodic  &  $0.22$M      &  \checkmark & &  \checkmark       &  $47.97$       &  $65.88$       &  $\bm{52.21}$  &  $\bm{71.82}$  \\
\textbf{Neg-Cosine}~\cite{Negative-cosine_ECCV20} & ECCV'20        & Conv64F      & Non-episodic  &  $0.22$M      &  \checkmark & &  \checkmark       &  $47.34$       &  $65.97$       &  $51.21$       &  $71.57$   \\
\textbf{SKD-GEN0}~\cite{rajasegaran2020self_SKD}  & BMVC'21        & Conv64F      & Non-episodic  &  $0.22$M      &  \checkmark & &  \checkmark       &  $48.14$       &  $\bm{66.36}$  &  $51.78$       &  $70.65$  \\
 \hdashline[1pt/1pt]
\textbf{MAML}~\cite{finn2017model}                & ICML'17        & Conv64F      & Meta         &  $0.12$M      &   & \checkmark  & \checkmark      &  $49.55$       &  $64.92$       &  $50.98$       &  $67.12$  \\
\textbf{Versa}~\cite{gordon2018versa}             & NeurIPS'18     & Conv64F      & Meta         &  $1.18$M      &   & \checkmark  & \checkmark      &  $52.75$       &  $67.40$       &  $52.28$       &  $69.41$  \\
\textbf{R2D2}~\cite{BertinettoHTV19_R2D2}         & ICLR'19        & Conv64F      & Meta         &  $0.11$M      &   & \checkmark  & \checkmark      &  $51.19$       &  $67.29$       &  $52.18$       &  $69.19$  \\
\textbf{LEO}~\cite{RusuRSVPOH19_LEO}              & ICLR'19        & Conv64F      & Meta         &  $1.20$M      &\checkmark &\checkmark &\checkmark &  $\bm{53.31}$  &  $\bm{67.47}$  &  $\bm{58.15}$  &  $\bm{74.21}$  \\
\textbf{MTL}~\cite{sun2019meta}                   & CVPR'19        & Conv64F      & Meta         &  $1.80$M      &\checkmark &\checkmark &\checkmark &  $46.70$       &  $64.79$       &  $49.11$       &  $69.13$  \\
\textbf{ANIL}~\cite{raghu2020ANIL}                & ICLR'20        & Conv64F      & Meta         &  $0.12$M      &  &\checkmark& \checkmark          &  $48.01$       &  $63.88$       &  $49.05$       &  $66.32$  \\
\textbf{BOIL}~\cite{OhYKY2021BOIL}                & ICLR'21        & Conv64F      & Meta         &  $0.12$M         &  & \checkmark   & \checkmark   &  $47.92$       &  $64.39$       &  $50.04$       &  $65.51$  \\
\hdashline[1pt/1pt]
\textbf{ProtoNet}~\cite{snell2017prototypical}    & NeurIPS'17     & Conv64F      & Metric       &  $0.11$M      &  &\checkmark &                    &  $47.05$       &  $68.56$       &  $46.11$       &  $70.07$  \\
\textbf{RelationNet}~\cite{sung2018learning}      & CVPR'18        & Conv64F      & Metric       &  $0.23$M      &  &\checkmark &                    &  $51.52$       &  $66.76$       &  $54.37$       &  $71.93$  \\
\textbf{CovaMNet}~\cite{li2019distribution}       & AAAI'19        & Conv64F      & Metric       &  $0.11$M      &  &\checkmark &                    &  $51.59$       &  $67.65$       &  $51.92$       &  $69.76$  \\
\textbf{DN4}~\cite{li2019revisiting}              & CVPR'19        & Conv64F      & Metric       &  $0.11$M      &  &\checkmark &                    &  $54.47$       &  $72.15$       &  $56.07$       &  $75.75$  \\
\textbf{CAN}~\cite{hou2019cross}                  & NeurIPS'19     & Conv64F      & Metric       &  $0.13$M      &\checkmark  &\checkmark  &         &  $55.88$       &  $70.98$       &  $55.96$       &  $70.52$  \\
\textbf{RENet}~\cite{kang2021renet}               & ICCV'21        & Conv64F      & Metric       &  $0.20$M      &\checkmark & \checkmark  &         &  $\bm{57.62}$  &  $\bm{74.14}$  &  $\bm{61.62}$  &  $\bm{76.74}$  \\
\midrule\midrule
\textbf{Baseline}~\cite{ChenLKWH19}               & ICLR'19        & ResNet12     & Non-episodic  &  $12.47$M    &  \checkmark &  &  \checkmark      &  $56.39$       &  $76.18$       &  $65.54$       &  $83.46$  \\
\textbf{Baseline++}~\cite{ChenLKWH19}             & ICML'19        & ResNet12     & Non-episodic  &  $12.47$M    &  \checkmark &  &  \checkmark      &  $58.79$       &  $75.31$       &  $66.32$       &  $83.05$  \\
\textbf{RFS-simple}~\cite{TianWKTI20_RFS}         & ECCV'20        & ResNet12     & Non-episodic  &  $12.47$M    &  \checkmark &  &  \checkmark      &  $61.65$       &  $78.88$       &  $70.55$       &  $84.74$  \\
\textbf{Neg-Cosine}~\cite{Negative-cosine_ECCV20} & ECCV'20        & ResNet12     & Non-episodic  &  $12.47$M    &  \checkmark &  &  \checkmark      &  $60.60$       &  $78.80$       &  $70.15$       &  $84.94$   \\
\textbf{SKD-GEN0}~\cite{rajasegaran2020self_SKD}  & BMVC'21        & ResNet12     & Non-episodic  &  $12.47$M    &  \checkmark &  &  \checkmark      &  $\bm{66.40}$  &  $\bm{83.06}$  &  $\bm{71.90}$  &  $\bm{86.20}$  \\
\hdashline[1pt/1pt]
\textbf{Versa}~\cite{gordon2018versa}             & NeurIPS'18     & ResNet12     & Meta         &  $13.25$M    & &\checkmark  &  \checkmark         &  $55.71$       &  $70.05$       &  $57.14$       &  $75.48$  \\
\textbf{R2D2}~\cite{BertinettoHTV19_R2D2}         & ICLR'19        & ResNet12     & Meta         &  $12.42$M    &  &\checkmark &  \checkmark         &  $59.52$       &  $74.61$       &  $65.07$       &  $83.04$  \\
\textbf{LEO}~\cite{RusuRSVPOH19_LEO}              & ICLR'19        & ResNet12     & Meta         &  $12.60$M    &\checkmark &\checkmark &\checkmark  &  $56.62$       &  $69.99$       &  $64.75$       &  $81.42$  \\
\textbf{MTL}~\cite{sun2019meta}                   & CVPR'19        & ResNet12     & Meta         & $13.13$M     &\checkmark &\checkmark &\checkmark  &  $\bm{62.67}$  &  $\bm{79.16}$  &  $\bm{68.68}$  &  $\bm{84.58}$  \\
\textbf{ANIL}~\cite{raghu2020ANIL}                & ICLR'20        & ResNet12     & Meta         &  $12.43$M    &  &\checkmark  &\checkmark          &  $52.77$       &  $68.11$       &  $55.65$       &  $73.52$  \\
\textbf{BOIL}~\cite{OhYKY2021BOIL}                & ICLR'21        & ResNet12     & Meta         &  $12.43$M    &  &\checkmark  & \checkmark         &  $58.87$       &  $72.88$       &  $64.66$       &  $80.38$  \\
\hdashline[1pt/1pt]
\textbf{ProtoNet}~\cite{snell2017prototypical}    & NeurIPS'17     & ResNet12     & Metric       &  $12.42$M    &   &\checkmark    &                &  $57.10$       &  $74.20$       &  $62.93$       &  $83.30$  \\
\textbf{RelationNet}~\cite{sung2018learning}      & CVPR'18        & ResNet12     & Metric       &  $23.53$M    &   &\checkmark    &                &  $55.22$       &  $69.25$       &  $56.86$       &  $74.66$  \\
\textbf{CovaMNet}~\cite{li2019distribution}       & AAAI'19        & ResNet12     & Metric       &  $12.42$M    &   &\checkmark    &                &  $54.69$       &  $70.72$       &  $56.03$       &  $75.21$  \\
\textbf{DN4}~\cite{li2019revisiting}              & CVPR'19        & ResNet12     & Metric       &  $12.42$M    &   &\checkmark    &                &  $59.14$       &  $75.26$       &  $64.41$       &  $82.59$  \\
\textbf{CAN}~\cite{hou2019cross}                  & NeurIPS'19     & ResNet12     & Metric       &  $12.65$M    &\checkmark &\checkmark    &        &  $62.68$       &  $78.36$       &  $\bm{70.46}$  &  $\bm{84.50}$  \\
\textbf{RENet}~\cite{kang2021renet}               & ICCV'21        & ResNet12     & Metric       &  $12.67$M    &\checkmark &\checkmark   &         &  $\bm{64.81}$  &  $\bm{79.90}$  &  $70.14$       &  $82.70$  \\
\midrule\midrule
\textbf{Baseline}~\cite{ChenLKWH19}               & ICLR'19        & ResNet18     & Non-episodic  & $11.20$M     &  \checkmark  &  &  \checkmark     &  $54.11$       &  $74.44$       &  $64.65$       &  $82.73$  \\
\textbf{Baseline++}~\cite{ChenLKWH19}             & ICML'19        & ResNet18     & Non-episodic  & $11.20$M     &  \checkmark  & &  \checkmark      &  $52.70$       &  $75.36$       &  $65.85$       &  $83.33$  \\
\textbf{RFS-simple}~\cite{TianWKTI20_RFS}         & ECCV'20        & ResNet18     & Non-episodic  & $11.20$M     &  \checkmark  &  &  \checkmark     &  $61.65$       &  $76.60$       &  $69.14$       &  $83.21$  \\
\textbf{Neg-Cosine}~\cite{Negative-cosine_ECCV20} & ECCV'20        & ResNet18     & Non-episodic  &$11.20$M      &  \checkmark  &   & \checkmark     &  $60.99$       &  $76.40$       &  $68.36$       &  $83.77$   \\
\textbf{SKD-GEN0}~\cite{rajasegaran2020self_SKD}  & BMVC'21        & ResNet18     & Non-episodic  & $11.20$M     &  \checkmark  &   &  \checkmark    &  $\bm{66.18}$  &  $\bm{82.21}$  &  $\bm{70.00}$  &  $\bm{84.70}$  \\
\hdashline[1pt/1pt]
\textbf{Versa}~\cite{gordon2018versa}            & NeurIPS'18     & ResNet18      & Meta         &  $11.96$M    & &\checkmark &\checkmark           &  $55.08$       &  $69.16$       &  $57.30$       &  $75.67$  \\
\textbf{R2D2}~\cite{BertinettoHTV19_R2D2}        & ICLR'19        & ResNet18      & Meta         &  $11.17$M    & &\checkmark &\checkmark           &  $58.36$       &  $75.69$       &  $64.73$       &  $\bm{83.40}$  \\
\textbf{LEO}~\cite{RusuRSVPOH19_LEO}             & ICLR'19        & ResNet18      & Meta         &  $11.31$M    &\checkmark &\checkmark &\checkmark &  $57.51$       &  $69.33$       &  $64.02$       &  $78.89$  \\
\textbf{MTL}~\cite{sun2019meta}                  & CVPR'19        & ResNet18      & Meta         &  $11.88$M    &\checkmark &\checkmark &\checkmark &  $\bm{60.29}$  &  $\bm{76.25}$  &  $\bm{65.12}$  &  $79.99$  \\
\textbf{ANIL}~\cite{raghu2020ANIL}               & ICLR'20        & ResNet18      & Meta         &  $11.17$M    &  &\checkmark   &\checkmark        &  $52.96$       &  $65.88$       &  $55.81$       &  $73.53$  \\
\textbf{BOIL}~\cite{OhYKY2021BOIL}               & ICLR'21        & ResNet18      & Meta         &  $11.17$M    &  &\checkmark   & \checkmark       &  $57.85$       &  $71.88$       &  $62.26$       &  $77.94$  \\
\hdashline[1pt/1pt]
\textbf{ProtoNet}~\cite{snell2017prototypical}   & NeurIPS'17     & ResNet18      & Metric       &  $11.17$M    &  &\checkmark  &                   &  $58.48$       &  $75.16$       &  $63.50$       &  $82.51$  \\
\textbf{RelationNet}~\cite{sung2018learning}     & CVPR'18        & ResNet18      & Metric       &  $18.29$M    &   &\checkmark          &          &  $53.98$       &  $71.27$       &  $60.80$       &  $77.94$  \\
\textbf{DN4}~\cite{li2019revisiting}             & CVPR'19        & ResNet18      & Metric       &  $11.17$M    &   &\checkmark      &              &  $57.30$       &  $74.23$       &  $64.83$       &  $82.77$  \\
\textbf{CAN}~\cite{hou2019cross}                 & NeurIPS'19     & ResNet18      & Metric       &  $11.35$M    &\checkmark & \checkmark    &       &  $62.33$       &  $77.12$       &  $\bm{71.70}$  &  $\bm{84.61}$  \\
\textbf{RENet}~\cite{kang2021renet}              & ICCV'21        & ResNet18      & Metric       & $11.52$M     &\checkmark &  \checkmark   &       &  $\bm{66.21}$  &  $\bm{81.20}$  &  $71.53$       &  $84.55$  \\
\bottomrule
\end{tabular} 
\label{Table2_SOTA}
\end{table*}

\section{Evaluation Settings and LibFewShot}
\textbf{Datasets.}
Our main experiments are conducted on two benchmark datasets, \emph{i.e.,} \textit{miniImageNet}~\cite{vinyals2016matching} and \textit{tieredImageNet}~\cite{RenTRSSTLZ18}. Moreover, we also evaluate the cross-domain generalization ability of each FSL method, three fine-grained benchmark datasets, \emph{i.e.,} \textit{Stanford Dogs}~\cite{khosla2011novel}, \textit{Stanford Cars}~\cite{krause20133d} and \textit{CUB Birds-200-2011}~\cite{wah2011caltech}. Following the literature, each data set is split into training (auxiliary), validation and test sets, respectively. The details can be seen in Table~\ref{libfewshot_datasets}. Note that all the images in the above datasets are resized to a resolution of $84\!\times\!84$.

\textbf{Backbone Architectures.}
Following the literature~\cite{snell2017prototypical,lee2019meta,ChenLKWH19}, we adopt three different embedding backbones from shallow to deep, \emph{i.e.,} \textit{Conv64F}, \textit{ResNet12} and \textit{ResNet18}. Specifically, Conv64F contains four convolutional blocks, each of which consists of a convolutional (Conv) layer, a batch-normalization (BN) layer, a ReLU/LeakyReLU layer and a max-pooling (MP) layer, where the numbers of filters of these blocks are $\{64,64,64,64\}$. ResNet12 consists of four residual blocks, each of which further contains three convolutional blocks (each is built as Conv-BN-ReLU-MP) along with a skip connection layer, where the numbers of filters of these blocks are $\{64,160,320,640\}$. ResNet18 is the standard architecture used in~\cite{he2016deep}. One important difference between ResNet12 and ResNet18 is that ResNet12 uses Dropblock~\cite{GhiasiLL18_Dropblock} in each residual block, while ResNet18 does not. Note that, in Table~\ref{Table1}, the numbers of filters of ResNet12$^\dagger$ are $\{64,96,128,256\}$; Conv64F$^\dagger$ has five Conv blocks; and Conv64F$^\ddagger$ uses additional low-level features.

\textbf{Evaluation Protocols.}
Following the prior works~\cite{vinyals2016matching,snell2017prototypical,sung2018learning}, in this paper, we control the evaluation setting for all methods, evaluate them on $600$ sampled tasks and repeat this process five times, \emph{i.e.,} a total of $3000$ tasks. The top-1 mean accuracy will be reported. In addition, early works~\cite{vinyals2016matching,snell2017prototypical,sung2018learning} generally use raw evaluation, \emph{i.e.,} directly resizing the test image into $84\times 84$ for evaluation, while the recent work~\cite{ChenLKWH19,ye2020few,yang2021free} has tried single center crop evaluation like the generic image classification~\cite{he2016deep}. To make the comparison more fair and can adapt to future development, we follow the latest setting to use the single center crop evaluation.

\textbf{Bag of Tricks.} Many works in other fields show that \textit{trick matters}, especially in deep learning~\cite{HeZ0ZXL19_bag_of_tricks,AT_bag_of_tricks}, so as to FSL. For example, some recent FSL works have introduced such ``tricks'', such as knowledge distillation~\cite{TianWKTI20_RFS,rajasegaran2020self_SKD}, self-supervision~\cite{gidaris2019boosting,SuMH20_selfSupervision} and Mixup~\cite{Mangla20_S2M2_mixup}, into the FSL problem. Specifically, we empirically summarize some key training tricks in FSL as below: (1) using data augmentation at the training stage (Train-DA); (2) augmenting the support set multiple times at the test stage (Test-DA); (3) pre-training on the auxiliary set (Pre-train); (4) global classification of using the global labels of the auxiliary set (Global-label); (5) Larger Episode size, \emph{i.e.,} increasing the number of tasks at each iteration; (6) higher way-number or higher shot-number during the episodic training; (7) using knowledge distillation or self-distillation (KD); (8) using self-supervision (SS); (9) using Dropblock; (10) using label smoothing; and (11) using more learnable parameters.

\textbf{LibFewShot.} In the literature, many implementation details or ``tricks'' are only briefly mentioned or even overlooked in many FSL works. However, these non-trivial tricks may lead to significant \textit{algorithm-agnostic performance boost}, which will make the comparison somewhat unfair and make some conclusions untenable. In this sense, it will not only make beginners struggle with the reproduction of other comparison methods, but also hinder them from developing their own methods. On the other hand, there are many interesting issues worth studying under a unified framework, including (1) the doubts on the necessity of meta- or episodic-training mechanism posted by the recent non-episodic based methods, (2) the effects of different deep learning tricks on the FSL problem, (3) the actual progress of FSL in the case of restricting deep learning tricks, (4) the effect of transformers on FSL.

Therefore, to address the above issues, we develop \textit{a comprehensive library for few-shot learning (LibFewShot)} by re-implementing the state-of-the-art methods into the same framework and applying the same training tricks to the maximum extent. So far, eighteen representative methods have been investigated, including five \textit{non-episodic based methods}, seven \textit{meta-learning based methods} and six \textit{metric-learning based methods}, where the details can be seen in Table~\ref{Table1}. The details of the architecture of LibFewShot are illustrated in the supplementary material.

\section{Experimental Results and Discussions}
\subsection{Reproduction Results} 
To validate the correctness of our re-implementation, we adopt the original settings of these methods and re-implement them with LibFewShot. Our re-implemented results and their originally reported results on \textit{mini}ImageNet are shown in Table~\ref{Table1}. Importantly, the original implementation details and tricks are also listed in detail, which can intuitively show the differences between different FSL methods on the implementation.

From Table~\ref{Table1}, we can observe that: (1) different FSL methods employ different backbones, especially the early meta- and metric-learning based methods only use a shallow backbone of Conv64F, making the comparison between different methods of using different backbones somewhat unfair; (2) non-episodic based methods, especially RFS~\cite{TianWKTI20_RFS} and SKD~\cite{rajasegaran2020self_SKD}, indeed obtain the state-of-the-art results, but they employ a much deeper backbone and much more tricks than other methods; (3) both knowledge distillation (KD) and self-supervision (SS) can significantly boost the performance; (4) when using additional pre-training or global classification with global labels, MTL~\cite{sun2019meta}, CAN~\cite{hou2019cross} and RENet~\cite{kang2021renet} can achieve much better results. Notably, RENet consistently outperforms RFS-simple and RFS-distill no matter the officially reported results or our re-implemented results in both the 1-shot and 5-shot settings, \textit{which shows great potential of the paradigm of pre-training $+$ meta-training}.

\subsection{Overview of the State of the Art} 
As seen in Table~\ref{Table1}, the implementation details of different methods vary a lot, which cannot trustingly reflect the actual progress of FSL. To this end, we keep most of the common training tricks consistent for all the methods except some special tricks (\emph{e.g.,}  self-supervision (SS) for SKD-GEN0~\cite{rajasegaran2020self_SKD}) to make a relatively fair comparison. To be specific, we use the exactly same embedding backbone, fixed input image size, \emph{i.e.,} $84\times 84$, the standard data augmentations (which consists of Resize, RandomCrop, RandomHorizonFlip and ColorJitter) during training, and center crop evaluation for all the methods. As for the optimizer and learning rate scheduler, only minor modifications are made according to their settings in the original papers. All methods are trained with 100 epochs, except for Versa with 200 epochs. Also, 2000 episodes are used for \emph{mini}ImageNet per epoch, while 5000 episodes for \emph{tiered}ImageNet. The results are reported in Table~\ref{Table2_SOTA}.

First, when using Conv64F as the embedding backbone, we can see that metric-learning based methods generally achieve relatively better results than meta-learning based and non-episodic based methods. This may be because when the embedding backbone is shallow, \emph{i.e.,} the feature representation is weak, the design or selection of metric function will be important. As also can be seen, the best methods of the metric-learning based and meta-learning based methods are RENet and LEO, respectively. One common characteristic of these two methods is that they employ both global- and local-label during training, which shows both global and local labels will benefit few-shot learning. In Section~\ref{Section_ablation}, we will further demonstrate this point. Similarly, MTL and CAN also employ such a kind of training strategy, both of which can obtain competitive results too. It is worth mentioning that some methods use more trainable parameters, such as Versa, LEO and MTL, which will also somewhat affect the final results.

Second, when using ResNet12 or ResNet18, a much deeper network, as the embedding backbone, we observe that the non-episodic based methods especially RFS-simple, Neg-Cosine and SKD-GEN0, perform significantly better than local-label based meta- and metric-learning methods, such as Versa, R2D2, ANIL, BOIL, ProtoNet, RelationNet, CovaMNet and DN4. However, when both global and local labels are used, we can see that MTL, CAN and RENet will be on par and even superior to RFS-simple and Neg-Cosine. Note that SKD-GEN0 utilizes an additional self-supervision trick, which essentially is not so fair for other methods. It is also worth noting that when using ResNet12 as the backbone most of these methods will perform much better than using ResNet18. This is because ResNet12 is much wider than ResNet18, \emph{i.e.,} ResNet12 enjoys more parameters than ResNet18. In addition, Dropblock is also used in ResNet12, while ResNet18 does not.

\begin{table*}[!tp]\footnotesize
	\centering
    \extrarowheight=3pt
    \tabcolsep=8pt
	\caption{\textbf{The necessity of test-tuning in the test phase}, where RFS-simple is taken as an example method by using different test settings. Test-tune means test-tuning of using the support set at the test stage. $\ell_2$ means using $\ell_2$ normalization. ($\mathcal{C}$, $\mathcal{K}$) means $\mathcal{C}$-way $\mathcal{K}$-shot tasks.}
	\begin{tabular}{@{}lccccccccccc@{}}
	\toprule
    \textbf{Method}     &\textbf{Classifier}  &\textbf{Test-tune}   &$\bm{\ell_2}$   &\textbf{(5,1)}   &\textbf{(5,5)}  &\textbf{(10,1)}    &\textbf{(10,5)}    &\textbf{(15,1)}      &\textbf{(15,5)}      &\textbf{(20,1)} &\textbf{(20,5)} \\
    \midrule
    \multirow{2}{*}{\textbf{RFS-simple}} &LR     &\Checkmark  &\XSolidBrush &$58.69$      &$77.72$       &$43.87$       &$64.86$      &$36.25$      &$57.08$       &$31.24$       &$51.70$  \\
                                         &LR     &\Checkmark  &\Checkmark   &$\bm{61.61}$ &$78.80$       &$\bm{45.85}$  &$65.18$      &$\bm{37.72}$ &$57.01$       &$\bm{32.47}$  &$51.25$  \\
    \hdashline[1pt/1pt]
    \multirow{2}{*}{\textbf{RFS-NN}}    &$1$-NN  &\XSolidBrush&\XSolidBrush &$56.74$      &$78.27$       &$41.64$       &$65.73$      &$33.87$      &$58.06$       &$29.19$       &$52.70$  \\
                                        &$1$-NN  &\XSolidBrush&\Checkmark   &$60.64$      &$\bm{78.85}$  &$44.89$       &$\bm{66.20}$ &$36.86$      &$\bm{58.46}$  &$31.70$       &$\bm{53.15}$  \\
    \bottomrule
	\end{tabular}
	\label{Table4_FT}
\end{table*}

\subsection{The Necessity of Episodic- and Meta-training}
\label{Section_4.3}
One concern of our work is to investigate the necessity of episodic- and meta-training and try to answer the questions raised in Section~\ref{Section_3}. To this end, we employ RFS-simple~\cite{TianWKTI20_RFS} without Test-DA as the benchmark, and use its pre-trained ResNet12 on \emph{mini}ImageNet as the embedding backbone.

\textbf{Is test-tuning really important at the test stage in FSL?} From Table~\ref{Table2_SOTA}, we can see that RFS-simple consistently outperforms Baseline and Baseline++ when using ResNet12 or ResNet18 as the embedding backbone. However, the only difference between them is that RFS adopts logistic regression (LR) with $\ell_2$ normalization as the classifier in the test phase. That is to say, for each novel test task, a new LR classifier will be specially learned during test (\emph{i.e.,} test-tuning). Instead of using LR, we directly employ $\ell_2$ normalization based $1$-NN for the final test classification, \emph{i.e.,} a non-parametric classifier without requiring test-tuning, and name this new variant as RFS-NN.

As seen in Table~\ref{Table4_FT}, without using $\ell_2$ normalization, the performance of RFS-simple will significantly degrade, especially in the 1-shot setting. In addition, using neither test-tuning nor $\ell_2$ normalization, RFS-NN still performs better than RFS-simple on the 5-shot tasks. More importantly, when only using $\ell_2$ normalization, RFS-NN could achieve very competitive results as RFS-simple on the 1-shot tasks and clearly outperforms RFS-simple on the 5-shot tasks especially in the higher way settings. This reveals that \textit{LR or test-tuning is not so important in the limited-data regime, but a good embedding and the $\ell_2$ normalization are!} One reason is that $\ell_2$ normalization can align the distributions between the source and target domains (base and novel classes) to some extent to guarantee good transferability (evidence can be seen in~\cite{xu2019larger}).

In summary, we argue that test-tuning is not so necessary at the test stage in FSL because of the limited-data regime. In another word, a few labeled examples normally cannot learn an effective classifier in the test phase. This also demonstrates that the paradigm of metric-learning based methods is reasonable.

\textbf{Is episodic- or meta-training necessary at the training stage in FSL?} To answer this question, we employ three methods, \emph{i.e.,} ProtoNet, DN4 and R2D2, to episodic/meta training (fine-tuning) the same pre-trained feature embedding used in RFS-simple. That is to say, this is a kind of paradigm of using pre-training with global labels and meta fine-tuning with local labels. Note that for ProtoNet, we use a cosine distance instead of the original Euclidean distance, because the above analysis shows that $\ell_2$ normalization is important. Specifically, for each method, we perform $\mathcal{C}$-way $1$-shot episodic/meta training and the corresponding $\mathcal{C}'$-way $1$-shot testing.

\begin{figure*}[!tp]
\centering
    \includegraphics[height=0.4\textheight]{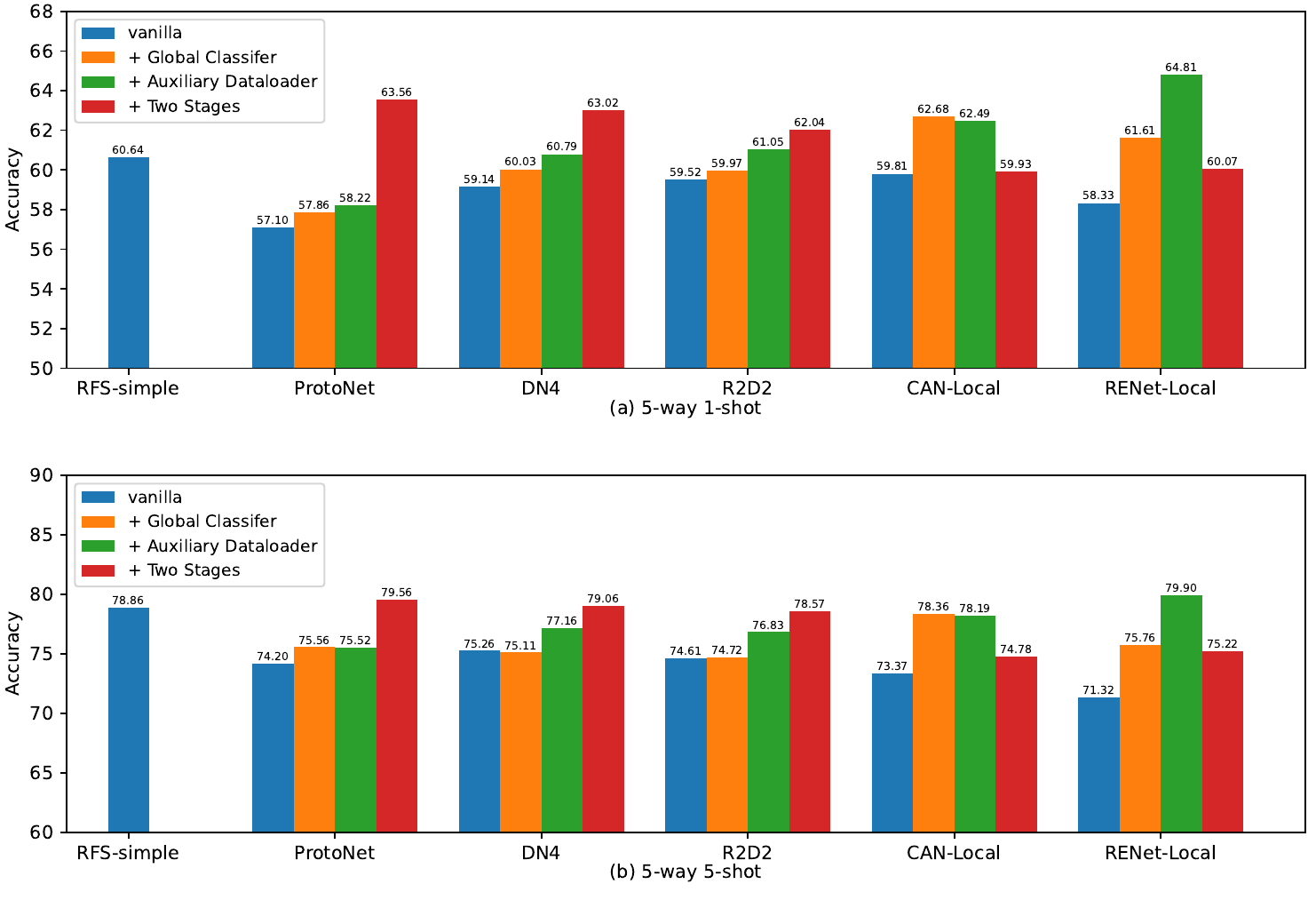}
\caption{\textbf{Effects of using both global and local labels} on \emph{mini}ImageNet with a ResNet12 backbone under both 5-way 1-shot and 5-shot settings.}
\label{Figure3}
\end{figure*}

The results are shown in Figure~\ref{Figure3}, whose details can be found in Section~\ref{Section_ablation}. As seen, for ProtoNet, DN4 and R2D2, all the episodic- or meta- fine-tuning (\emph{i.e.,} Two Stages) can further improve the performance over RFS-simple. Note that the data augmentation used in the fine-tuning phase (which only contains Resize and RandomResizedCrop) shall be somewhat different from the data augmentation used in the pre-training phase, which is one indetectable trick for this success overlooked in the literature. We also notice that if using somewhat different or more complicated metric functions in the meta fine-tuning phase will also significantly benefits the final performance. Overall, this simple experiment demonstrates that pre-training may obtain a good initial embedding, but it is not the optimal one. \textit{Therefore, we argue that the episodic- or meta-training is worthy of further investigation at the training stage in FSL.} 

\subsection{Ablation Study on Non-trivial Tricks} 
\label{Section_ablation}
In this section, we select multiple representative methods from the three kinds of FSL methods and conduct ablation studies on multiple non-trivial tricks. To be specific, \textit{mini}ImageNet and ResNet12 are taken as the default benchmark dataset and embedding backbone, respectively. Also, both 5-way 1-shot and 5-shot tasks are considered. Typically, non-trivial tricks, including \textit{global-label}, \textit{local-label}, \textit{strong data augmentation}, \textit{knowledge distillation}, \textit{label smoothing} and \textit{self supervision}, are taken into consideration. In addition, RFS-simple, R2D2, ProtoNet, DN4, CAN-Local and RENet-Local will be selected as the representative FSL methods. Note that CAN-Local and RENet-Local are the variants of CAN and RENet of only using the local labels, respectively.

\subsubsection{Effects of Global and Local Labels} To thoroughly investigate the effects of global and local labels, we consider three types of how to use both of them: (1) Pre-training with global labels at the first stage and episodic fine-tuning with local labels using a specific method at the second stage, we name this type as \textit{Two Stages} for simplicity; (2) Global classifier with global labels + local classifier with local labels through episodic-training in a multi-task learning manner, we name this type as \textit{Global Classifier} for short; (3) is similar to (2) but uses different data loaders for these two tasks, which is first adopted in RENet. We name the type of (3) as \textit{Auxiliary Dataloader} for simplicity. Note that both (2) and (3) are one-stage methods, which use both global and local labels via a single training process. From the results in Fig.~\ref{Figure3}, we have the following observations.

(1) All three types of using both global and local labels work well, and can effectively improve the performance over the vanilla version of only using the local or global labels; For example, under the 5-way 1-shot setting, \textit{ProtoNet + Two Stages} (using both global and local labels) can achieve an accuracy of $63.56\%$, obtaining $6.46\%$ improvements over the vanilla version (using local labels only), which is significantly better than RFS-simple ($60.64\%$). Similarly, \textit{DN4 + Two Stages} ($63.02\%$) and \textit{R2D2 + Two Stages} ($62.04\%$) gains $4.34\%$ and $2.52\%$ improvements over their vanilla versions, respectively.

(2) For the episodic-training based methods of only using local labels, such as ProtoNet, DN4 and R2D2, using extra global labels by pre-training (\emph{i.e.,} a two-stage way) is the most effective way. In contrast, for the methods originally designed with both global and local labels, such as CAN and RENet, using the global labels with a multi-task manner (\emph{i.e.,} a one-stage way) is generally the more effective way.

\begin{figure}[!tp]
\centering
\includegraphics[height=0.25\textheight]{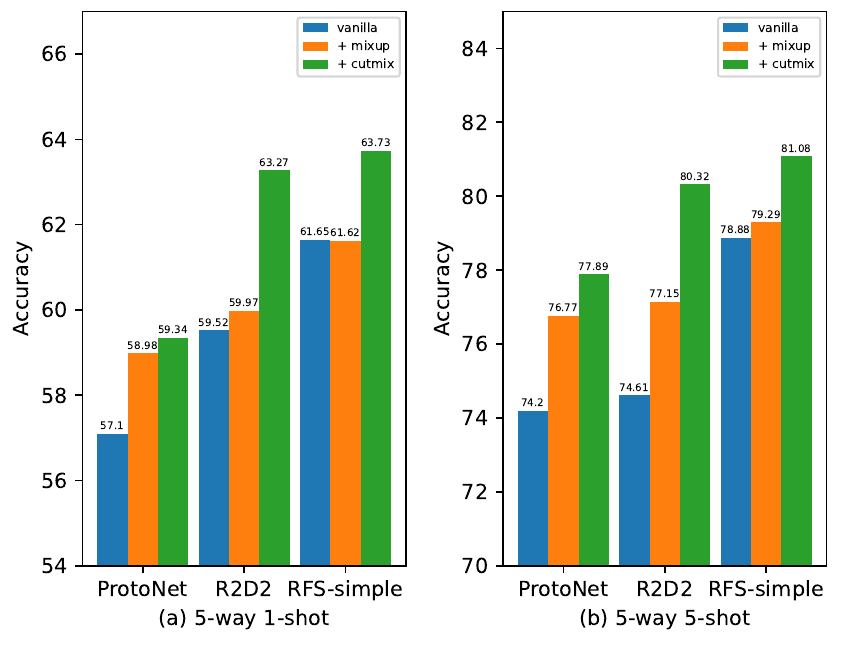}
\caption{\textbf{Effects of using strong data augmentation}, \emph{i.e.,} mixup and cutmix, for ProtoNet, R2D2 and RFS-simple on \emph{mini}ImageNet with a ResNet12 backbone.}
\label{Figure4}
\end{figure}

\subsubsection{Effect of Strong Data Augmentation} Data augmentation is one of the most general and effective tricks in the field of deep learning. Therefore, it will be interesting to investigate the effect of strong data augmentation techniques for FSL. To be specific, we take mixup~\cite{mixup_ICLR18} with alpha of $0.2$ and cutmix~\cite{CutMix_ICCV19} with alpha of $1.0$ (alpha is the hyper-parameter of the beta distribution in these two tricks) as the representative strong data augmentations and take ProtoNet, R2D2 and RFS-simple as the representative FSL methods.

From the results in Fig.~\ref{Figure4}, we can observe that: (1) cutmix can significantly improve the performance of all the three FSL methods. For example, in the 1-shot setting, cutmix obtains $2.24\%$, $3.75\%$, and $2.08\%$ improvements over the vanilla versions of ProtoNet, R2D2 and RFS-simple, respectively. Similarly, in the 5-shot setting, cutmix obtains $3.69\%$, $5.71\%$, and $2.2\%$ improvements over the three vanilla versions, respectively.
(2) In most cases, mixup can effectively improve the performance of all the three FSL methods, especially for ProtoNet and R2D2. For example, in the 5-shot setting, mixup gains $2.57\%$, $2.54\%$, $0.41\%$ improvements over the vanilla versions of ProtoNet, R2D2 and RFS-simple, respectively. 
(3) When using strong data augmentation, \emph{e.g.,} cutmix, R2D2 is surprisingly competitive to RFS-simple. That is to say, by just using a simple data augmentation technique, \textit{R2D2 + cutmix} achieves accuracies of $63.27\%$ and $80.32\%$ in the 1-shot and 5-shot settings, respectively, which are already very competitive in Table~\ref{Table2_SOTA}.

\begin{figure}[!tp]
\centering
\includegraphics[height=0.25\textheight]{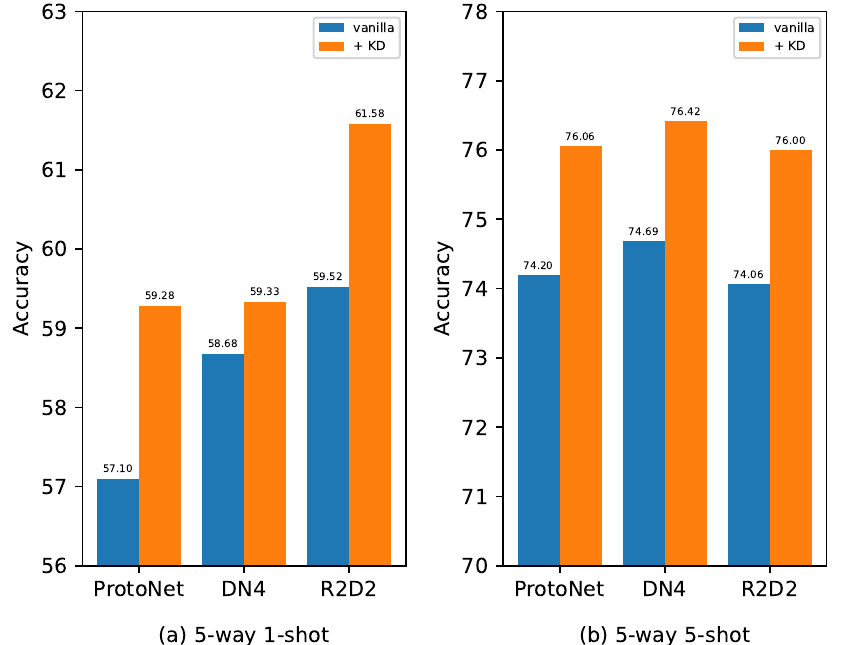}
\caption{\textbf{Effect of knowledge distillation (KD)} for ProtoNet, DN4 and R2D2 on \emph{mini}ImageNet with a ResNet12 backbone.}
\label{Figure5}
\end{figure}

\subsubsection{Effect of Knowledge Distillation} 
Knowledge distillation (KD) has been introduced for FSL by some recent FSL methods, such as RFS-distill and SKD-GEN1. To further verify its effect, we apply KD to three early FSL methods, \emph{i.e.,} ProtoNet, DN4 and R2D2. Following RFS-distill and SKD-GEN1, we also adopt the self-distillation to distill the knowledge from a trained few-shot model to a new identical model initialized from scratch for once.

As seen in Fig.~\ref{Figure5}, in both settings of 1-shot and 5-shot, KD can consistently improve the performance of the vanilla versions of all the three FSL methods. For example, under the 5-shot setting, the gained improvements for ProtoNet, DN4 and R2D2 are $1.86\%$, $1.73\%$, $1.94\%$, respectively. Similarly, under the 1-shot setting, KD boosts the performance of ProtoNet, DN4 and R2D2 by $2.18\%$, $0.65\%$, $2.06\%$ improvements, respectively. This shows that KD is indeed effective in FSL. On the other hand, we shall notice that KD is a general trick, which is applicable to other FSL methods.

\subsubsection{Effect of Self Supervision} 
Self-supervision methods have been shown to be effective in many fields of deep learning with the advantage of requiring no additional annotation costs. In the field of FSL, Gudarus \emph{et al.}~\cite{Boosting_SSL_GidarisBKPC19} has used the additional rotation prediction self-supervision~\cite{RotNet_ICLR18} as an auxiliary parallel task. Similarly, SKD~\cite{rajasegaran2020self_SKD} also adopts the rotation prediction as an auxiliary task but in the pre-training stage. These works have demonstrated that self-supervision, especially rotation prediction self-supervision, is effective for FSL.

\begin{figure}[!tp]
\centering
\includegraphics[height=0.25\textheight]{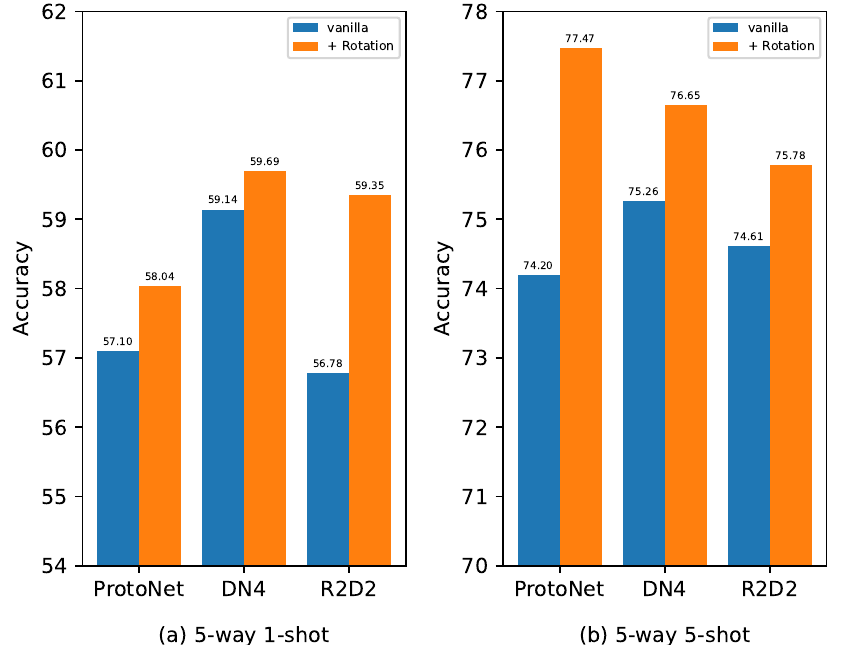}
\caption{\textbf{Effect of self supervision (SS)} for ProtoNet, DN4 and R2D2 on \emph{mini}ImageNet with a ResNet12 backbone.}
\label{Figure_SS}
\end{figure}

To further investigate whether self supervision is a general trick for other FSL methods, we take the rotation prediction as an auxiliary task and employ ProtoNet, DN4 and R2D2 as three representative FSL methods. Specifically, all methods are trained from scratch through the episodic-training mechanism by using an additional rotation classifier in a multi-task manner. The results are shown in Fig.~\ref{Figure_SS}. We can see that no matter in the 5-way 1-shot setting nor 5-way 5-shot setting, the rotation prediction self supervision can consistently boost the performance of all three FSL methods. For example, in the 1-shot setting, the rotation self-supervision can boost ProtoNet, DN4 and R2D2 for $0.94\%$, $0.55\%$, $2.57\%$ improvements, respectively. Similarly, in the 5-shot setting, the gained performance improvements of ProtoNet, DN4 and R2D2 are $3.27\%$, $1.39\%$, $1.17\%$, respectively. This reveals that the self-supervision task, especially the rotation prediction task, is indeed effective for FSL and can be a general trick for different FSL methods.

\begin{figure}[!tp]
\centering
\includegraphics[height=0.25\textheight]{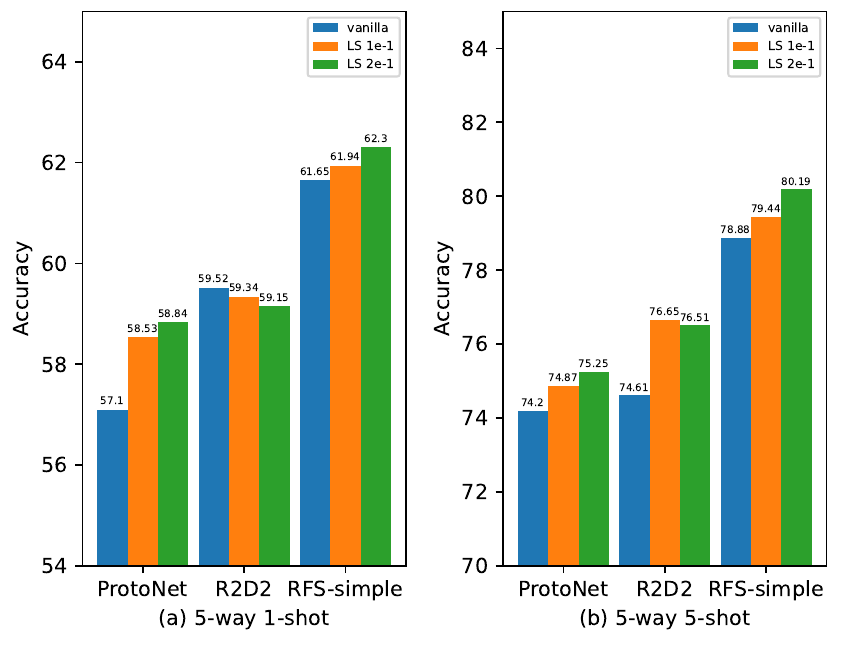}
\caption{\textbf{Effect of using label smoothing (LS)} for ProtoNet, R2D2 and RFS-simple on \emph{mini}ImageNet with ResNet12.}
\label{Figure6}
\end{figure}

\subsubsection{Effect of Label Smoothing} Label smoothing (LS)~\cite{Inceptionv2_CVPR16} attempts to prevent the model to be over-confident by softening the ground-truth labels, which is a common trick in the field of deep learning and representation learning. Therefore, it is also interesting to investigate its effect on the problem of FSL. To be specific, we apply LS to three FSL methods, including ProtoNet, R2D2 and RFS-simple. In addition, for each method of using LS, we adopt two different hyper-parameters for LS, \emph{i.e.,} $0.1$ and $0.2$, to balance the weight between the original ground-truth and the uniform distribution.

The results are reported in Fig.~\ref{Figure6}. From the results, not surprisingly, we can see that in most cases, LS can further improve the performance of the vanilla versions of the three FSL methods, especially in the 5-shot setting. For example, in the 5-shot setting, LS (0.2) boosts the performance of ProtoNet, R2D2 and RFS-simple by $1.05\%$, $1.90\%$, $1.31\%$ improvements, respectively. This reveals that label smoothing can also be a general trick for boosting the performance of FSL methods.

\begin{table*}[!t]
   \centering
    \extrarowheight=4pt
    \tabcolsep=5pt
	\caption{\textbf{Using Transformer in FSL} on both \textit{mini}ImageNet and \textit{tiered}ImageNet. All model are trained from scratch without pre-training. The fourth column shows the total number of trainable parameters used by each model and the fifth column denotes the details of optimization during training.}
    \begin{tabular}{ccccccccc}
    \toprule
    \multirow{2}{*}{\textbf{Method}} &\multirow{2}{*}{\textbf{Embedding}} &\multirow{2}{*}{\textbf{Image Size}}
    &\multirow{2}{*}{\textbf{Parameters}}
    &\multirow{2}{*}{\textbf{Optimizer} / \textbf{Lr} / \textbf{Decay}}  &\multicolumn{2}{c}{\textbf{\textit{mini}ImageNet}} &\multicolumn{2}{c}{\textbf{\textit{tiered}ImageNet}} \\
    \cmidrule{6-9}
     & & & &     &(5,1) &(5,5) &(5,1) &(5,5) \\
   \midrule
   \multirow{2}{*}{ProtoNet}  & ResNet12 &$84\times 84$ &12.42M &Adam/1e-3/Step   & 57.10 & 74.20 & 62.93 & 83.30 \\
                              & Swin-T   &$84\times 84$ &26.59M &AdamW/5e-4/Cosine & 57.23 & 74.67 & 62.33 & 81.83 \\
   \midrule
   \multirow{2}{*}{DN4}       & ResNet12 &$84\times 84$ &12.42M &Adam/1e-3/Step   & 59.14 & 75.26 & 64.41 & 82.59\\
                              & Swin-T   &$84\times 84$ &26.59M &AdamW/1e-3/Cosine & 54.29 & 67.46 & 64.94 & 79.36 \\
   \midrule
   \multirow{2}{*}{R2D2}      & ResNet12 &$84\times 84$ &12.42M &Adam/1e-3/Step   & 59.52 & 74.61 & 65.07 & 83.04 \\
                              & Swin-T   &$84\times 84$ &26.59M &AdamW/5e-5/Cosine & 50.41 & 61.68 & 54.71 & 67.68  \\
   \midrule
   \multirow{2}{*}{RFS}       & ResNet12 &$84\times 84$ &12.47M &Adam/1e-3/Step    & 61.65 & 78.88 & 70.55 & 84.74 \\
                              & Swin-T   &$84\times 84$ &26.64M &AdamW/5e-4/Cosine & 58.13 & 76.94 & 68.12 & 83.99  \\
\bottomrule
\end{tabular}
\label{tab:transformer}
\end{table*}

\begin{table*}[!tp]\footnotesize
	\centering
    \extrarowheight=4pt
    \tabcolsep=1pt
	\caption{\textbf{Cross-domain transferability}. All methods are learned from the source domain (\emph{e.g.,} \emph{mini}ImageNet), and directly evaluated on the test set of the target domain (\emph{i.e.,} Stanford Dogs, Stanford Cars and CUB Birds-200-2011) with a ResNet12 backbone.}
	\begin{tabular}{@{}lccccc|cc|cc|cc|cc|cc|cc@{}}
	\toprule
    \multirow{2}{*}{\textbf{Method}} &\multirow{2}{*}{\textbf{Type}} &\multirow{2}{*}{\textbf{Test-tune}} &\multirow{2}{*}{\textbf{Global-label}} &\multirow{2}{*}{\textbf{KD}} &\multirow{2}{*}{\textbf{SS}} &\multicolumn{2}{c|}{\emph{\textbf{mini}}$\rightarrow$\textbf{Dogs}}  
    &\multicolumn{2}{c|}{\emph{\textbf{mini}}$\rightarrow$\textbf{Birds}} &\multicolumn{2}{c|}{\emph{\textbf{mini}}$\rightarrow$\textbf{Cars}} 
                                                  &\multicolumn{2}{c|}{\emph{\textbf{tiered}}$\rightarrow$\textbf{Dogs}} &\multicolumn{2}{c|}{\emph{\textbf{tiered}}$\rightarrow$\textbf{Birds}} &\multicolumn{2}{c}{\emph{\textbf{tiered}}$\rightarrow$\textbf{Cars}}  \\
                            & & & & &                    &$1$-shot  &$5$-shot  &$1$-shot  &$5$-shot &$1$-shot  &$5$-shot  &$1$-shot  &$5$-shot  &$1$-shot  &$5$-shot &$1$-shot  &$5$-shot\\
    \midrule
    \textbf{Baseline}~\cite{ChenLKWH19}         &  Non-episodic  &  \checkmark  &\checkmark &   &  &$51.21$  &$69.54$  &$47.81$  &$68.79$ &$33.07$  &$\bm{50.72}$  &  $76.82$  &  $92.18$  &  $68.32$  &  $87.34$ &  $34.88$  &  $50.72$  \\
    \textbf{Baseline++}~\cite{ChenLKWH19}       &  Non-episodic  &  \checkmark  &\checkmark &   &  &$51.16$  &$65.62$  &$41.34$  &$57.60$ &$27.52$  &$37.17$  &  $79.53$  &  $92.13$  &  $67.60$  &  $84.33$ &  $33.30$  &  $50.29$  \\
    \textbf{RFS-simple}~\cite{TianWKTI20_RFS}   &  Non-episodic  &  \checkmark  &\checkmark &   &  &  $56.15$  &  $71.90$  &  $47.88$  &  $65.53$ &  $32.18$  &  $44.80$  &  $81.46$  &  $92.85$  &  $70.96$  &  $85.89$   &  $34.17$  &  $48.58$ \\
    \textbf{RFS-distill}~\cite{TianWKTI20_RFS}  &  Non-episodic &\checkmark&\checkmark &\checkmark &&$\bm{58.46}$  &$\bm{74.88}$  &$50.12$  &$69.19$ &$33.20$  &$47.83$  &$81.44$  &$93.01$  &$\bm{71.35}$  &$\bm{87.42}$  &$\bm{35.61}$  &  $\bm{53.36}$  \\
    \textbf{Neg-Cosine}~\cite{Negative-cosine_ECCV20} &Non-episodic  &\checkmark &\checkmark &   &  &$56.09$ &$72.31$    &$47.39$    &$66.34$  &$31.30$ &  $45.12$  &  $\bm{81.87}$  &  $\bm{93.25}$  &  $70.53$  &  $86.70$  &  $33.31$      &  $49.80$  \\
    \textbf{SKD-GEN0}~\cite{rajasegaran2020self_SKD}  &Non-episodic  &\checkmark &\checkmark & &\checkmark  & $56.29$  &  $74.51$  &  $\bm{51.94}$  &  $72.71$  &  $33.04$  &  $48.66$  &  $73.55$  &  $89.18$  &  $70.32$  &  $87.06$  &  $35.03$  &  $50.87$  \\
    \textbf{SKD-GEN1}~\cite{rajasegaran2020self_SKD}  &Non-episodic  &\checkmark &\checkmark &\checkmark &\checkmark  &$57.38$  &$75.20$  &$52.57$  &$\bm{73.11}$  &$\bm{33.39}$  &  $49.30$  &  $74.28$  &  $71.24$  &  $71.24$  &  $86.96$  &  $34.81$  &  $50.22$  \\
    \hdashline[1pt/1pt]
    \textbf{Versa}~\cite{gordon2018versa}             &  Meta         &\checkmark  & &          &       &  $42.35$    &  $57.83$    &  $40.98$    &  ${58.10}$  &  ${25.98}$  &  ${31.63}$  &  ${51.28}$  &  ${76.17}$  &  ${49.71}$  &  ${66.85}$  &  ${25.09}$  &  ${35.38}$  \\
    \textbf{R2D2}~\cite{BertinettoHTV19_R2D2}         &  Meta         &\checkmark  & &          &       &  ${52.16}$  &  ${68.72}$  &  ${44.87}$  &  ${62.47}$  &  ${29.37}$  &  ${45.42}$  &  ${65.58}$  &  ${87.62}$  &  ${59.81}$  & ${82.29}$   &  ${30.99}$  &  ${50.58}$  \\
    \textbf{LEO}~\cite{RusuRSVPOH19_LEO}              &  Meta         &\checkmark  &\checkmark  &   &   &  ${47.47}$  &  ${61.62}$  &  ${37.71}$  &  ${51.70}$  &  ${27.66}$  &  ${34.07}$  &  ${70.62}$  &  ${88.79}$  &  ${61.17}$  & ${72.05}$   &  ${29.34}$  &  ${35.76}$  \\
    \textbf{ANIL}~\cite{raghu2020ANIL}                &  Meta         &\checkmark  & &          &       &  ${31.41}$  &  ${50.31}$  &  ${35.16}$  &  ${49.78}$  &  ${27.56}$  &  ${32.94}$  &  ${43.08}$  &  ${69.22}$  &  ${41.67}$  &  ${63.98}$  &  ${28.78}$  &  ${38.67}$  \\
    \textbf{BOIL}~\cite{OhYKY2021BOIL}                &  Meta         &\checkmark  & &          &       &  ${44.60}$  &  ${60.75}$  &  ${44.81}$  &  ${59.68}$  &  ${28.77}$  &  ${37.65}$  &  ${63.50}$  &  ${83.54}$  &  ${60.22}$  &  ${78.38}$  &  ${31.44}$  &  ${46.91}$  \\
    \textbf{MTL}~\cite{sun2019meta}                   &  Meta         &\checkmark  &\checkmark & &      &  $\bm{54.35}$ &$\bm{72.11}$  &$\bm{48.01}$  &$\bm{66.39}$ &$\bm{31.97}$  &$\bm{46.77}$  &$\bm{72.23}$  &$\bm{89.41}$  &$\bm{66.66}$  &$\bm{85.49}$  &$\bm{35.20}$  &$\bm{53.94}$  \\
    \hdashline[1pt/1pt]
    \textbf{ProtoNet}~\cite{snell2017prototypical}    & Metric        &     &  &         &              &  ${40.62}$  &  ${65.80}$  &  ${44.17}$  &  $\bm{67.73}$  &  ${27.82}$  &  ${41.53}$  &  ${57.21}$  &  ${86.70}$  &  ${56.15}$  &  ${82.47}$  &  ${29.93}$  &  ${48.51}$  \\
    \textbf{RelationNet}~\cite{sung2018learning}      & Metric        &     &  &         &              &  ${38.77}$  &  ${57.96}$  &  ${40.45}$  &  ${53.49}$  &  ${26.69}$  &  ${31.70}$  &  $49.40$  &  $72.44$  &  $50.84$  &  $68.26$  &  $27.91$  &  $37.91$  \\
    \textbf{CovaMNet}~\cite{li2019distribution}       & Metric        &     &  &         &              &  ${40.18}$  &  ${53.30}$  &  ${40.24}$  &  ${48.66}$  &  ${28.96}$  &  ${33.37}$  &  ${46.75}$  &  ${65.06}$  &  ${36.04}$  &  ${58.88}$  &  ${29.10}$  &  ${36.43}$  \\
    \textbf{DN4}~\cite{li2019revisiting}              & Metric        &     &  &         &              &  ${43.72}$  &  ${59.82}$  &  ${42.77}$  &  ${61.73}$  &  ${29.08}$  &  ${43.66}$  &  ${55.36}$  &  ${76.83}$  &  ${55.78}$  &  ${74.84}$  &  ${32.81}$  &  ${47.71}$  \\
    \textbf{CAN}~\cite{hou2019cross}                  & Metric  &     &\checkmark  &         &          &  $\bm{56.03}$ &${71.34}$  &  ${43.94}$  &  ${62.37}$  &  ${29.09}$  &  ${39.16}$  &  ${76.56}$  &  ${90.57}$ &  $\bm{69.74}$  &  $\bm{85.95}$  &  $\bm{32.99}$  &  $\bm{48.73}$   \\
    \textbf{RENet}~\cite{kang2021renet}               & Metric  &     &\checkmark  &         &          &  ${53.60}$  &$\bm{71.44}$  &  $\bm{48.69}$  &  ${65.79}$  &  $\bm{31.09}$  &  $\bm{44.45}$  &  $\bm{77.67}$  &  $\bm{90.68}$  &  ${69.50}$  &  ${85.17}$  &  ${32.35}$  &  ${44.00}$  \\
    \bottomrule
	\end{tabular}
	\label{Tabel_cross_domain}
\end{table*}

\section{Using Transformer in FSL}
In recent years, many new vision transformer based models have been proposed, such as ViT~\cite{VIT_ICLR2021}, DeiT~\cite{DeiT_ICML2021}, DETR~\cite{DETR_ICLR2021}, SETR~\cite{SETR_CVPR2021}, and Swin-Transformer~\cite{SwinT_ICCV2021}, and shown promising results on a variety of vision tasks, compared with CNN architecture-based models. Therefore, it will be interesting to explore the effect of transformer in few-shot learning.

In particular, we implement four representative FSL methods, including ProtoNet, DN4, R2D2 and RFS, by using the tiny version of Swin Transformer (Swin-T for short)~\cite{SwinT_ICCV2021}, a powerful transformer network, as the embedding backbone. Also, the versions using ResNet12 as the embedding backbone for these methods are taken as comparisons. Note that the common Swin-T is designed for the generic image classification with the image size of $224\times 224$, while FSL normally uses an image size of $84\times84$. Therefore, for a fair comparison, we use an image size of $84\times84$ for Swin-T and change the down-sampling factors for each stage (linear embedding + transformer block) from $(4, 2, 2, 2)$ to $(3, 2, 2 ,1)$ but keep all the other architecture hyper-parameters by default. In terms of data augmentation, RandomCrop, RandomHorizontalFlip, and ColorJitter, which are commonly used in few-shot learning, could not achieve good results when using Swin-T as the backbone. Therefore, we add RandAugment, mixup and cutmix into the data augmentation transformations for better results. In addition, we find that transformer is sensitive to the learning rate, so we follow the literature of generic classification, use AdamW as the optimizer, and carefully adjust each model's learning rate.

The results are reported in Table~\ref{tab:transformer}. From the results, we observe that: (1) In most cases, Swin-T shows worse results than ResNet12 on both \emph{mini}ImageNet and \emph{tiered}ImageNet. This is because transformers generally rely on a large-scale dataset for training to achieve excellent performance, because transformers lack locality and translation invariance properties that existed in the CNN architecture~\cite{Cvt_ICCV2021,cao2022training}. (2) Swin-T can benefit from a large dataset, \emph{i.e.,} \emph{tiered}ImageNet. As seen, all methods especially DN4 and RFS, obtain much higher results on \emph{tiered}ImageNet than the results on \emph{mini}ImageNet under both 1-shot and 5-shot settings. (3) Interestingly, when using Swin-T, ProtoNet can achieve very competitive results with the results of using ResNet12. Also, DN4 with Swin-T achieves a higher result than DN4 with ResNet12 on \emph{tiered}ImageNet under the 1-shot setting.

In summary, the above analyses reveal that transformers, as the latest popular and powerful architecture, are also promising and showing good potential in the field of few-shot learning. However, the key to achieving this expectation is how to effectively address the generic limitation, \emph{i.e.,} the data-hungry property, of transformers.

\subsection{Cross-domain Transferability}
Cross-domain few-shot tasks have been introduced in many FSL works in the literature. Therefore, it will also be interesting to further evaluate the cross-domain transfer ability of different FSL methods in LibFewShot, which are not specially designed for this purpose. To this end, following the literature~\cite{ChenLKWH19}, we conduct an experiment on six cross-domain scenarios, \textit{e.g.,} \emph{mini}ImageNet$\rightarrow$Stanford Dogs (\emph{mini}$\rightarrow$Dogs for short). In this experiment, all the few-shot models are trained on the source domain, \textit{e.g.,} \emph{mini}ImageNet or \emph{tiered}ImageNet, using a ResNet12 backbone with the same setting in Table~\ref{Table2_SOTA} and directly tested on the target domain, \textit{e.g.,} Standford Dogs or CUB Birds.

The results are reported in Table~\ref{Tabel_cross_domain}. From the results, we have the following observations: (1) The models trained on \emph{mini}ImageNet and \emph{tiered}ImageNet can easily generalize to Stanford Dogs and CUB Birds that enjoy a small domain-shift. However, their performance significantly drops when performing on the Stanford Cars with a large domain-shift. (2) In most cases, especially in the large domain-shift scenario, \emph{i.e.,} \emph{mini}$\rightarrow$Cars, non-episodic based FSL methods perform somewhat better than metric-based or meta-based FSL methods. This means that pre-training is beneficial to cross-domain scenarios. Pre-training can easily bring prior strong representations which work well in natural image cross-domain scenarios. Especially, MTL, CAN and RENet use a pre-trained model or global labels, and they can also perform well on the \emph{mini}$\rightarrow$Cars task.  (3) We can see that the classification accuracy on the cross-domain target datasets is significantly lower than that on the in-domain target datasets. 
This reveals that the current state-of-the-art FSL methods cannot handle the cross-domain scenarios well, which needs to be further investigated in the future.





\section{Conclusions}
In this paper, we present \textit{a comprehensive library for few-shot learning} (LibFewShot) by re-implementing the state-of-the-art FSL methods in a unified framework. Through LibFewShot, first, we are able to make a relatively fair comparison between different methods to reflect the actual progress of FSL. Second, we emphasize and demonstrate the necessity of episodic- or meta-training. Third, we find that test-tuning is not very important at the test stage because of the limited-data setting in FSL, while a good embedding and $\ell_2$ normalization are truly important. Finally, we verify that many deep learning tricks are indeed non-trivial but are universal for different FSL methods. Also, we show that transformers are promising for FSL but are still needed to be further investigated in the future. We hope our work will facilitate healthy research on few-shot learning.

\section*{Acknowledgments}
This work is supported in part by the Science and Technology Innovation 2030 New Generation Artificial Intelligence Major Project (2021ZD0113303), National Natural Science Foundation of China (62106100, 62192783, 61806092), Jiangsu Natural Science Foundation (BK20221441), Collaborative Innovation Center of Novel Software Technology and Industrialization, and Jiangsu Provincial Double-Innovation Doctor Program (JSSCBS20210021).





\ifCLASSOPTIONcaptionsoff
  \newpage
\fi



%

\bibliographystyle{IEEEtran}
\bibliography{egbib}

\begin{thebibliography}{10}
\providecommand{\url}[1]{#1}
\csname url@samestyle\endcsname
\providecommand{\newblock}{\relax}
\providecommand{\bibinfo}[2]{#2}
\providecommand{\BIBentrySTDinterwordspacing}{\spaceskip=0pt\relax}
\providecommand{\BIBentryALTinterwordstretchfactor}{4}
\providecommand{\BIBentryALTinterwordspacing}{\spaceskip=\fontdimen2\font plus
\BIBentryALTinterwordstretchfactor\fontdimen3\font minus
  \fontdimen4\font\relax}
\providecommand{\BIBforeignlanguage}[2]{{%
\expandafter\ifx\csname l@#1\endcsname\relax
\typeout{** WARNING: IEEEtran.bst: No hyphenation pattern has been}%
\typeout{** loaded for the language `#1'. Using the pattern for}%
\typeout{** the default language instead.}%
\else
\language=\csname l@#1\endcsname
\fi
#2}}
\providecommand{\BIBdecl}{\relax}
\BIBdecl

\bibitem{fei2006one}
L.~Fei-Fei, R.~Fergus, and P.~Perona, ``One-shot learning of object
  categories,'' \emph{IEEE transactions on pattern analysis and machine
  intelligence}, vol.~28, no.~4, pp. 594--611, 2006.

\bibitem{vinyals2016matching}
O.~Vinyals, C.~Blundell, T.~Lillicrap, D.~Wierstra \emph{et~al.}, ``Matching
  networks for one shot learning,'' in \emph{Proceedings of the Conference on
  Neural Information Processing Systems (NeurIPS)}, 2016, pp. 3630--3638.

\bibitem{finn2017model}
C.~Finn, P.~Abbeel, and S.~Levine, ``Model-agnostic meta-learning for fast
  adaptation of deep networks,'' in \emph{Proceedings of the International
  Conference on Machine Learning (ICML)}, 2017, pp. 1126--1135.

\bibitem{sung2018learning}
F.~Sung, Y.~Yang, L.~Zhang, T.~Xiang, P.~H. Torr, and T.~M. Hospedales,
  ``Learning to compare: Relation network for few-shot learning,'' in
  \emph{Proceedings of the IEEE Conference on Computer Vision and Pattern
  Recognition (CVPR)}, 2018, pp. 1199--1208.

\bibitem{sun2020meta}
Q.~Sun, Y.~Liu, Z.~Chen, T.-S. Chua, and B.~Schiele, ``Meta-transfer learning
  through hard tasks,'' \emph{IEEE Transactions on Pattern Analysis and Machine
  Intelligence}, 2020.

\bibitem{simon2020dsn}
C.~Simon, P.~Koniusz, R.~Nock, and M.~Harandi, ``Adaptive subspaces for
  few-shot learning,'' in \emph{Proceedings of the IEEE Conference on Computer
  Vision and Pattern Recognition (CVPR)}, 2020, pp. 4135--4144.

\bibitem{LiWHSGL20_ADM}
W.~Li, L.~Wang, J.~Huo, Y.~Shi, Y.~Gao, and J.~Luo, ``Asymmetric distribution
  measure for few-shot learning,'' in \emph{Proceedings of the International
  Joint Conference on Artificial Intelligence (IJCAI)}, 2020, pp. 2957--2963.

\bibitem{ye2020heterogeneous}
H.-J. Ye, D.-C. Zhan, Y.~Jiang, and Z.-H. Zhou, ``Heterogeneous few-shot model
  rectification with semantic mapping,'' \emph{IEEE Transactions on Pattern
  Analysis and Machine Intelligence}, 2020.

\bibitem{yang2021free}
S.~Yang, L.~Liu, and M.~Xu, ``Free lunch for few-shot learning: Distribution
  calibration,'' in \emph{Proceedings of the International Conference on
  Learning Representations (ICLR)}, 2021.

\bibitem{abbas2022sharp}
M.~Abbas, Q.~Xiao, L.~Chen, P.-Y. Chen, and T.~Chen, ``Sharp-maml:
  Sharpness-aware model-agnostic meta learning,'' in \emph{Proceedings of the
  International Conference on Machine Learning (ICML)}, 2022.

\bibitem{gidaris2018dynamic}
S.~Gidaris and N.~Komodakis, ``Dynamic few-shot visual learning without
  forgetting,'' in \emph{Proceedings of the IEEE Conference on Computer Vision
  and Pattern Recognition (CVPR)}, 2018, pp. 4367--4375.

\bibitem{lee2019meta}
K.~Lee, S.~Maji, A.~Ravichandran, and S.~Soatto, ``Meta-learning with
  differentiable convex optimization,'' in \emph{Proceedings of the IEEE
  Conference on Computer Vision and Pattern Recognition (CVPR)}, 2019, pp.
  10\,657--10\,665.

\bibitem{sun2019meta}
Q.~Sun, Y.~Liu, T.-S. Chua, and B.~Schiele, ``Meta-transfer learning for
  few-shot learning,'' in \emph{Proceedings of the IEEE Conference on Computer
  Vision and Pattern Recognition (CVPR)}, 2019, pp. 403--412.

\bibitem{li2019distribution}
W.~Li, J.~Xu, J.~Huo, L.~Wang, Y.~Gao, and J.~Luo, ``Distribution consistency
  based covariance metric networks for few-shot learning,'' in
  \emph{Proceedings of the AAAI Conference on Artificial Intelligence (AAAI)},
  2019, pp. 8642--8649.

\bibitem{li2019revisiting}
W.~Li, L.~Wang, J.~Xu, J.~Huo, Y.~Gao, and J.~Luo, ``Revisiting local
  descriptor based image-to-class measure for few-shot learning,'' in
  \emph{Proceedings of the IEEE Conference on Computer Vision and Pattern
  Recognition (CVPR)}, 2019, pp. 7260--7268.

\bibitem{ye2020few}
H.-J. Ye, H.~Hu, D.-C. Zhan, and F.~Sha, ``Few-shot learning via embedding
  adaptation with set-to-set functions,'' in \emph{Proceedings of the IEEE
  Conference on Computer Vision and Pattern Recognition (CVPR)}, 2020, pp.
  8808--8817.

\bibitem{xie2022joint}
J.~Xie, F.~Long, J.~Lv, Q.~Wang, and P.~Li, ``Joint distribution matters: Deep
  brownian distance covariance for few-shot classification,'' in
  \emph{Proceedings of the {IEEE} Conference on Computer Vision and Pattern
  Recognition (CVPR)}, 2022, pp. 7972--7981.

\bibitem{afrasiyabi2022matching}
A.~Afrasiyabi, H.~Larochelle, J.-F. Lalonde, and C.~Gagn{\'e}, ``Matching
  feature sets for few-shot image classification,'' in \emph{Proceedings of the
  {IEEE} Conference on Computer Vision and Pattern Recognition (CVPR)}, 2022,
  pp. 9014--9024.

\bibitem{pan2009survey}
S.~J. Pan and Q.~Yang, ``A survey on transfer learning,'' \emph{IEEE
  Transactions on Knowledge and Data Engineering}, pp. 1345--1359, 2009.

\bibitem{santoro2016meta}
A.~Santoro, S.~Bartunov, M.~Botvinick, D.~Wierstra, and T.~Lillicrap,
  ``Meta-learning with memory-augmented neural networks,'' in \emph{Proceedings
  of the International Conference on Machine Learning (ICML)}, 2016, pp.
  1842--1850.

\bibitem{ravi2017optimization}
S.~Ravi and H.~Larochelle, ``Optimization as a model for few-shot learning,''
  in \emph{Proceedings of the International Conference on Learning
  Representations (ICLR)}, 2017.

\bibitem{koch2015siamese}
G.~Koch, R.~Zemel, and R.~Salakhutdinov, ``Siamese neural networks for one-shot
  image recognition,'' in \emph{Proceedings of the International Conference on
  Machine Learning (ICML) Deep Learning Workshop}, 2015.

\bibitem{snell2017prototypical}
J.~Snell, K.~Swersky, R.~Zemel, and R.~Zemel, ``Prototypical networks for
  few-shot learning,'' in \emph{Proceedings of the Conference on Neural
  Information Processing Systems (NeurIPS)}, 2017, pp. 4077--4087.

\bibitem{thrun1998lifelong}
S.~Thrun, ``Lifelong learning algorithms,'' in \emph{Learning to Learn}, 1998,
  pp. 181--209.

\bibitem{vilalta2002perspective}
R.~Vilalta and Y.~Drissi, ``A perspective view and survey of meta-learning,''
  \emph{Artificial Intelligence Review}, pp. 77--95, 2002.

\bibitem{BertinettoHTV19_R2D2}
L.~Bertinetto, J.~F. Henriques, P.~H.~S. Torr, and A.~Vedaldi, ``Meta-learning
  with differentiable closed-form solvers,'' in \emph{Proceedings of the
  International Conference on Learning Representations (ICLR)}, 2019.

\bibitem{ChenLKWH19}
W.~Chen, Y.~Liu, Z.~Kira, Y.~F. Wang, and J.~Huang, ``A closer look at few-shot
  classification,'' in \emph{Proceedings of the International Conference on
  Learning Representations (ICLR)}, 2019.

\bibitem{dhillon2020baseline}
G.~S. Dhillon, P.~Chaudhari, A.~Ravichandran, and S.~Soatto, ``A baseline for
  few-shot image classification,'' in \emph{Proceedings of the International
  Conference on Learning Representations (ICLR)}, 2020.

\bibitem{chen2020new}
Y.~Chen, X.~Wang, Z.~Liu, H.~Xu, and T.~Darrell, ``A new meta-baseline for
  few-shot learning,'' \emph{arXiv preprint arXiv:2003.04390}, 2020.

\bibitem{TianWKTI20_RFS}
Y.~Tian, Y.~Wang, D.~Krishnan, J.~B. Tenenbaum, and P.~Isola, ``Rethinking
  few-shot image classification: {A} good embedding is all you need?'' in
  \emph{Proceedings of the European Conference on Computer Vision (ECCV)},
  2020, pp. 266--282.

\bibitem{thrun1998learning}
S.~Thrun and L.~Pratt, ``Learning to learn: Introduction and overview,'' in
  \emph{Learning to Learn}, 1998, pp. 3--17.

\bibitem{rajasegaran2020self_SKD}
J.~Rajasegaran, S.~Khan, M.~Hayat, F.~S. Khan, and M.~Shah, ``Self-supervised
  knowledge distillation for few-shot learning,'' in \emph{Proceedings of the
  British Machine Vision Conference (BMVC)}, 2021.

\bibitem{Mangla20_S2M2_mixup}
P.~Mangla, M.~Singh, A.~Sinha, N.~Kumari, V.~N. Balasubramanian, and
  B.~Krishnamurthy, ``Charting the right manifold: Manifold mixup for few-shot
  learning,'' in \emph{Proceedings of the IEEE Winter Conference on
  Applications of Computer Vision (WACV)}, 2020, pp. 2207--2216.

\bibitem{Negative-cosine_ECCV20}
B.~Liu, Y.~Cao, Y.~Lin, Q.~Li, Z.~Zhang, M.~Long, and H.~Hu, ``Negative margin
  matters: Understanding margin in few-shot classification,'' in
  \emph{Proceedings of the European Conference on Computer Vision (ECCV)}, vol.
  12349, 2020, pp. 438--455.

\bibitem{rotation_2018}
S.~Gidaris, P.~Singh, and N.~Komodakis, ``Unsupervised representation learning
  by predicting image rotations,'' in \emph{Proceedings of the International
  Conference on Learning Representations (ICLR)}, 2018.

\bibitem{hinton2015_KD}
G.~Hinton, O.~Vinyals, and J.~Dean, ``Distilling the knowledge in a neural
  network,'' in \emph{Proceedings of the Conference on Neural Information
  Processing Systems (NeurIPS) Deep Learning and Representation Learning
  Workshop}, 2015.

\bibitem{manifoldMixup2019}
V.~Verma, A.~Lamb, C.~Beckham, A.~Najafi, I.~Mitliagkas, D.~Lopez{-}Paz, and
  Y.~Bengio, ``Manifold mixup: Better representations by interpolating hidden
  states,'' in \emph{Proceedings of the International Conference on Machine
  Learning (ICML)}, vol.~97, 2019, pp. 6438--6447.

\bibitem{Exemplar-2014}
A.~Dosovitskiy, J.~T. Springenberg, M.~A. Riedmiller, and T.~Brox,
  ``Discriminative unsupervised feature learning with convolutional neural
  networks,'' in \emph{Proceedings of the Conference on Neural Information
  Processing Systems (NeurIPS)}, 2014, pp. 766--774.

\bibitem{gordon2018versa}
J.~Gordon, J.~Bronskill, M.~Bauer, S.~Nowozin, and R.~E. Turner, ``Versa:
  Versatile and efficient few-shot learning,'' in \emph{Proceedings of the
  Conference on Neural Information Processing Systems (NeurIPS)}, 2018, pp.
  1--9.

\bibitem{RusuRSVPOH19_LEO}
A.~A. Rusu, D.~Rao, J.~Sygnowski, O.~Vinyals, R.~Pascanu, S.~Osindero, and
  R.~Hadsell, ``Meta-learning with latent embedding optimization,'' in
  \emph{Proceedings of the International Conference on Learning Representations
  (ICLR)}, 2019.

\bibitem{raghu2020ANIL}
A.~Raghu, M.~Raghu, S.~Bengio, and O.~Vinyals, ``Rapid learning or feature
  reuse? towards understanding the effectiveness of maml,'' in
  \emph{Proceedings of the International Conference on Learning Representations
  (ICLR)}, 2020.

\bibitem{xu2020attentional}
W.~Xu, H.~Wang, Z.~Tu \emph{et~al.}, ``Attentional constellation nets for
  few-shot learning,'' in \emph{Proceedings of the International Conference on
  Learning Representations (ICLR)}, 2020.

\bibitem{OhYKY2021BOIL}
J.~Oh, H.~Yoo, C.~Kim, and S.~Yun, ``{BOIL:} towards representation change for
  few-shot learning,'' in \emph{Proceedings of the International Conference on
  Learning Representations (ICLR)}, 2021.

\bibitem{DoerschGZ20}
C.~Doersch, A.~Gupta, and A.~Zisserman, ``Crosstransformers: Spatially-aware
  few-shot transfer,'' in \emph{Proceedings of the Conference on Neural
  Information Processing Systems (NeurIPS)}, 2020.

\bibitem{zhang2020deepemd}
C.~Zhang, Y.~Cai, G.~Lin, and C.~Shen, ``Deepemd: Few-shot image classification
  with differentiable earth mover's distance and structured classifiers,'' in
  \emph{Proceedings of the IEEE Conference on Computer Vision and Pattern
  Recognition (CVPR)}, 2020, pp. 12\,203--12\,213.

\bibitem{wertheimer2021few}
D.~Wertheimer, L.~Tang, and B.~Hariharan, ``Few-shot classification with
  feature map reconstruction networks,'' in \emph{Proceedings of the IEEE
  Conference on Computer Vision and Pattern Recognition (CVPR)}, 2021, pp.
  8012--8021.

\bibitem{kang2021relational}
D.~Kang, H.~Kwon, J.~Min, and M.~Cho, ``Relational embedding for few-shot
  classification,'' in \emph{Proceedings of the IEEE International Conference
  on Computer Vision (ICCV)}, 2021.

\bibitem{hou2019cross}
R.~Hou, H.~Chang, B.~Ma, S.~Shan, and X.~Chen, ``Cross attention network for
  few-shot classification,'' in \emph{Proceedings of the Conference on Neural
  Information Processing Systems (NeurIPS)}, 2019.

\bibitem{kang2021renet}
D.~Kang, H.~Kwon, J.~Min, and M.~Cho, ``Relational embedding for few-shot
  classification,'' in \emph{Proceedings of the IEEE International Conference
  on Computer Vision (ICCV)}, 2021.

\bibitem{RenTRSSTLZ18}
M.~Ren, E.~Triantafillou, S.~Ravi, J.~Snell, K.~Swersky, J.~B. Tenenbaum,
  H.~Larochelle, and R.~S. Zemel, ``Meta-learning for semi-supervised few-shot
  classification,'' in \emph{Proceedings of the International Conference on
  Learning Representations (ICLR)}, 2018.

\bibitem{khosla2011novel}
A.~Khosla, N.~Jayadevaprakash, B.~Yao, and F.-F. Li, ``Novel dataset for
  fine-grained image categorization: Stanford dogs,'' in \emph{Proceedings of
  the IEEE Conference on Computer Vision and Pattern Recognition (CVPR)
  Workshop}, 2011, p.~1.

\bibitem{krause20133d}
J.~Krause, M.~Stark, J.~Deng, and L.~Fei-Fei, ``3d object representations for
  fine-grained categorization,'' in \emph{Proceedings of the IEEE International
  Conference on Computer Vision (ICCV) Workshop}, 2013, pp. 554--561.

\bibitem{wah2011caltech}
C.~Wah, S.~Branson, P.~Welinder, P.~Perona, and S.~Belongie, ``The caltech-ucsd
  birds-200-2011 dataset,'' 2011.

\bibitem{he2016deep}
K.~He, X.~Zhang, S.~Ren, and J.~Sun, ``Deep residual learning for image
  recognition,'' in \emph{Proceedings of the IEEE Conference on Computer Vision
  and Pattern Recognition (CVPR)}, 2016, pp. 770--778.

\bibitem{GhiasiLL18_Dropblock}
G.~Ghiasi, T.~Lin, and Q.~V. Le, ``Dropblock: {A} regularization method for
  convolutional networks,'' in \emph{Proceedings of the Conference on Neural
  Information Processing Systems (NeurIPS)}, 2018, pp. 10\,750--10\,760.

\bibitem{HeZ0ZXL19_bag_of_tricks}
T.~He, Z.~Zhang, H.~Zhang, Z.~Zhang, J.~Xie, and M.~Li, ``Bag of tricks for
  image classification with convolutional neural networks,'' in
  \emph{Proceedings of the IEEE Conference on Computer Vision and Pattern
  Recognition (CVPR)}, 2019, pp. 558--567.

\bibitem{AT_bag_of_tricks}
T.~Pang, X.~Yang, Y.~Dong, H.~Su, and J.~Zhu, ``Bag of tricks for adversarial
  training,'' in \emph{Proceedings of the International Conference on Learning
  Representations (ICLR)}, 2021.

\bibitem{gidaris2019boosting}
S.~Gidaris, A.~Bursuc, N.~Komodakis, P.~P{\'e}rez, and M.~Cord, ``Boosting
  few-shot visual learning with self-supervision,'' in \emph{Proceedings of the
  IEEE International Conference on Computer Vision (ICCV)}, 2019, pp.
  8059--8068.

\bibitem{SuMH20_selfSupervision}
J.~Su, S.~Maji, and B.~Hariharan, ``When does self-supervision improve few-shot
  learning?'' in \emph{Proceedings of the European Conference on Computer
  Vision (ECCV)}, 2020, pp. 645--666.

\bibitem{xu2019larger}
R.~Xu, G.~Li, J.~Yang, and L.~Lin, ``Larger norm more transferable: An adaptive
  feature norm approach for unsupervised domain adaptation,'' in
  \emph{Proceedings of the IEEE International Conference on Computer Vision
  (ICCV)}, 2019, pp. 1426--1435.

\bibitem{mixup_ICLR18}
H.~Zhang, M.~Ciss{\'{e}}, Y.~N. Dauphin, and D.~Lopez{-}Paz, ``mixup: Beyond
  empirical risk minimization,'' in \emph{Proceedings of the International
  Conference on Learning Representations (ICLR)}, 2018.

\bibitem{CutMix_ICCV19}
S.~Yun, D.~Han, S.~Chun, S.~J. Oh, Y.~Yoo, and J.~Choe, ``Cutmix:
  Regularization strategy to train strong classifiers with localizable
  features,'' in \emph{Proceedings of the IEEE International Conference on
  Computer Vision (ICCV)}, 2019, pp. 6022--6031.

\bibitem{Boosting_SSL_GidarisBKPC19}
S.~Gidaris, A.~Bursuc, N.~Komodakis, P.~P{\'{e}}rez, and M.~Cord, ``Boosting
  few-shot visual learning with self-supervision,'' in \emph{Proceedings of the
  {IEEE/CVF} International Conference on Computer Vision (ICCV)}, 2019, pp.
  8058--8067.

\bibitem{RotNet_ICLR18}
S.~Gidaris, P.~Singh, and N.~Komodakis, ``Unsupervised representation learning
  by predicting image rotations,'' in \emph{Proceedings of the International
  Conference on Learning Representations (ICLR)}, 2018.

\bibitem{Inceptionv2_CVPR16}
C.~Szegedy, V.~Vanhoucke, S.~Ioffe, J.~Shlens, and Z.~Wojna, ``Rethinking the
  inception architecture for computer vision,'' in \emph{Proceedings of the
  IEEE Conference on Computer Vision and Pattern Recognition (CVPR)}, 2016, pp.
  2818--2826.

\bibitem{VIT_ICLR2021}
A.~Dosovitskiy, L.~Beyer, A.~Kolesnikov, D.~Weissenborn, X.~Zhai,
  T.~Unterthiner, M.~Dehghani, M.~Minderer, G.~Heigold, S.~Gelly, J.~Uszkoreit,
  and N.~Houlsby, ``An image is worth 16x16 words: Transformers for image
  recognition at scale,'' in \emph{Proceedings of the International Conference
  on Learning Representations (ICLR)}, 2021.

\bibitem{DeiT_ICML2021}
H.~Touvron, M.~Cord, M.~Douze, F.~Massa, A.~Sablayrolles, and H.~J{\'{e}}gou,
  ``Training data-efficient image transformers {\&} distillation through
  attention,'' in \emph{Proceedings of the International Conference on Machine
  Learning (ICML)}, vol. 139, 2021, pp. 10\,347--10\,357.

\bibitem{DETR_ICLR2021}
X.~Zhu, W.~Su, L.~Lu, B.~Li, X.~Wang, and J.~Dai, ``Deformable {DETR:}
  deformable transformers for end-to-end object detection,'' in
  \emph{Proceedings of the International Conference on Learning Representations
  (ICLR)}, 2021.

\bibitem{SETR_CVPR2021}
S.~Zheng, J.~Lu, H.~Zhao, X.~Zhu, Z.~Luo, Y.~Wang, Y.~Fu, J.~Feng, T.~Xiang,
  P.~H.~S. Torr, and L.~Zhang, ``Rethinking semantic segmentation from a
  sequence-to-sequence perspective with transformers,'' in \emph{Proceedings of
  the {IEEE} Conference on Computer Vision and Pattern Recognition (CVPR)},
  2021, pp. 6881--6890.

\bibitem{SwinT_ICCV2021}
Z.~Liu, Y.~Lin, Y.~Cao, H.~Hu, Y.~Wei, Z.~Zhang, S.~Lin, and B.~Guo, ``Swin
  transformer: Hierarchical vision transformer using shifted windows,'' in
  \emph{Proceedings of the {IEEE/CVF} International Conference on Computer
  Vision (ICCV)}, 2021, pp. 9992--10\,002.

\bibitem{Cvt_ICCV2021}
H.~Wu, B.~Xiao, N.~Codella, M.~Liu, X.~Dai, L.~Yuan, and L.~Zhang, ``Cvt:
  Introducing convolutions to vision transformers,'' in \emph{Proceedings of
  the {IEEE/CVF} International Conference on Computer Vision (ICCV)}, 2021, pp.
  22--31.

\bibitem{cao2022training}
Y.-H. Cao, H.~Yu, and J.~Wu, ``Training vision transformers with only 2040
  images,'' \emph{arXiv preprint arXiv:2201.10728}, 2022.

\bibitem{cubuk2019autoaugment}
E.~D. Cubuk, B.~Zoph, D.~Mane, V.~Vasudevan, and Q.~V. Le, ``Autoaugment:
  Learning augmentation policies from data,'' \emph{arXiv preprint
  arXiv:1805.09501}, 2019.

\bibitem{devries2017improved}
T.~DeVries and G.~W. Taylor, ``Improved regularization of convolutional neural
  networks with cutout,'' \emph{arXiv preprint arXiv:1708.04552}, 2017.

\bibitem{cubuk2019randaugment}
E.~D. Cubuk, B.~Zoph, J.~Shlens, and Q.~Le, ``Randaugment: Practical automated
  data augmentation with a reduced search space,'' in \emph{Proceedings of the
  Conference on Neural Information Processing Systems (NeurIPS)}, 2020.

\end{thebibliography}

\appendices
\section{Architecture}
\begin{figure}[!t]
\centering
\includegraphics[width=0.38\textwidth]{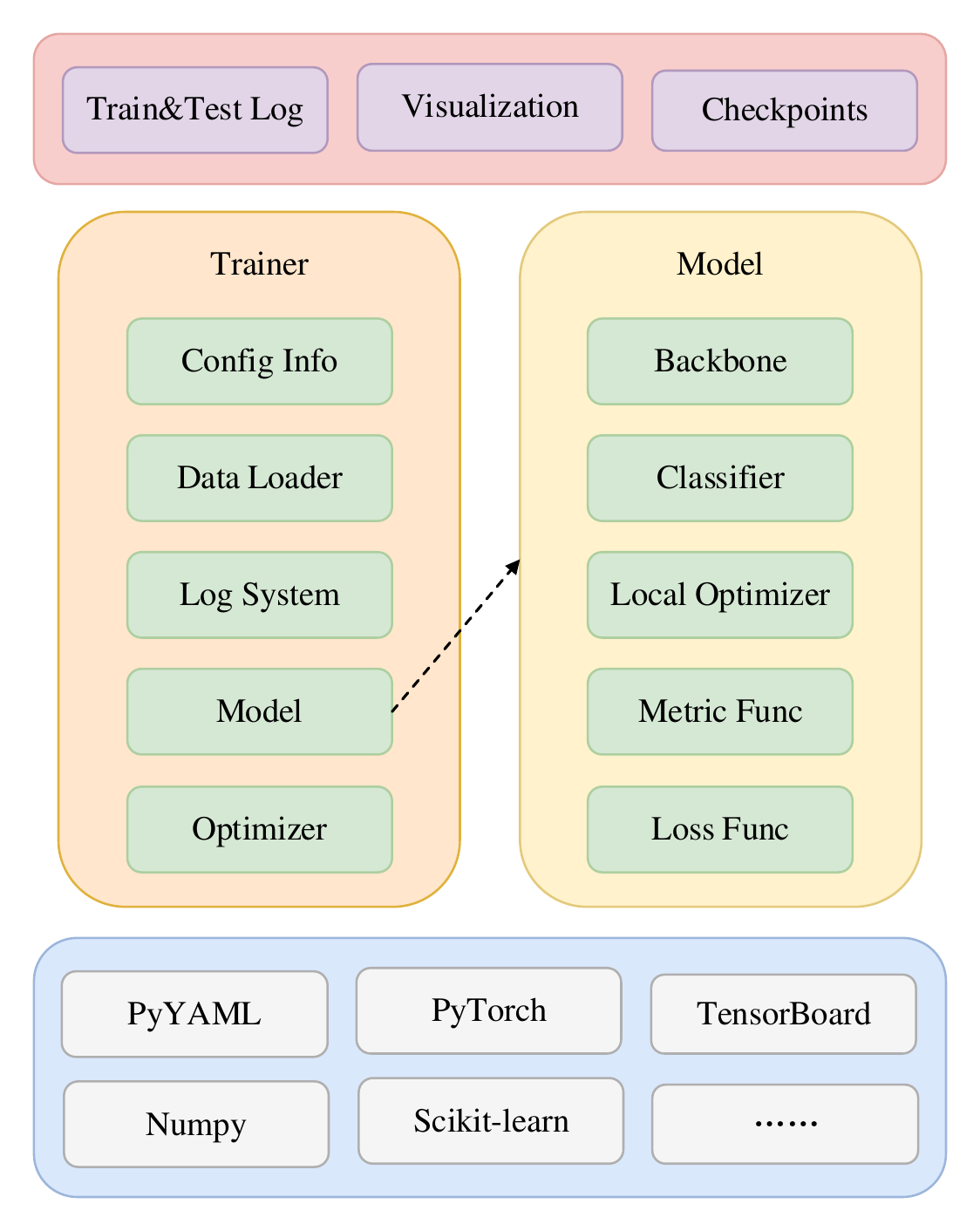}
\caption{Architecture of the proposed LibFewShot built on PyTorch.}
\label{framework_figure}
\end{figure}

LibFewShot is built on PyTorch~$1.5.0$, and its architecture can be seen in Fig.~\ref{framework_figure}. Because of the great differences between different FSL methods in terms of the network architecture, loss function and optimizer, it is difficult to directly integrate the existing FSL methods into the same framework. To address this issue, we disassemble each FSL method into multiple small common modules, aiming to integrate them into the same framework in a more flexible way. The details will be described in the following sections.

\subsection{Model}
The \textbf{Model} module belonging to the \textbf{Trainer} module is a key part of the whole framework, because all the network architectures of the FSL methods are implemented within this module. Specifically, \textbf{Model} consists of \textit{Backbone}, \textit{Classifier}, \textit{Local Optimizer}, \textit{Metric Function} and \textit{Loss Function}. Also, we will briefly describe some of these core parts.

\textbf{Backbone.}
The embedding backbone plays an important role in the field of deep learning. In order to support different requirements, LibFewShot provides options of the commonly used embedding modules, \textit{e.g.}, Conv64F, ResNet12, ResNet18, Wide ResNet (WRN) and Vision Transformer (ViT). Moreover, because some methods may need to modify the backbones in some cases, such as feature flattening, global average pooling and using multi-level features, LibFewShot can conveniently meet such kinds of requirements by simply modifying the configuration files.

\textbf{Classifier.} Despite the backbone part, the classifier may be the soul of one FSL method. However, we find that some FSL methods have inconsistent operations in the training phase and evaluation phase. To overcome this issue, we implement two functions in the \textit{Classifier} module, \textit{i.e.,} \textit{set\_forward\_loss function} and \textit{set\_forward function}, which can be flexibly used for the training mode and evaluation mode, respectively. As mentioned in the main paper, we divide the FSL methods into three categories, \textit{i.e.,} \textit{non-episodic based methods}, \textit{meta-learning based methods} and \textit{metric-learning based methods}. To avoid the duplication of work, we provide a category-dependent function for each category, and they all inherit the same \textit{abstract function}, in which the commonly used and model-agnostic hyper-parameters can be defined.

In addition, we notice that the official implementations of some FSL methods can only support single-task episodic training (\textit{i.e.,} one task in each mini-batch), which may make the FSL models be sensitive to the hyper-parameters and initializations. In contrast, some other FSL methods have already supported multi-task episodic training (\textit{i.e.,} multiple tasks in each mini-batch). To make a more fair comparison, we re-implement the architectures of the classifiers of the methods that can only support single-task episodic training to support multi-task episodic training. In this sense, users can realize this operation by simply modifying the parameter of \textit{episode size} in the configuration file.

\subsection{Dataloader}
LibFewShot provides a special dataloader, which can fulfill the requirements in few-shot learning. Specifically, LibFewShot assumes that all datasets have a similar file structure. It means that each dataset should have an image folder containing all the images and three csv files (train, validation and test) to indicate the image path and its corresponding class label. Moreover, LibFewShot will provide the processed datasets or the corresponding conversion program, depending on the open source protocol of these datasets.

In FSL, the input data structure is generally different from that of the generic computer vision tasks. In another word, in FSL, the smallest data unit is not an image but a task, which contains $5\times(5+15)=100$ images in a $5$-way $5$-shot setting when there are $15$ query images per class. For this reason, many open source codes for FSL only sample a whole task per thread, which severely limits the efficiency of data loading. Differently, LibFewShot deigns a \textit{CategoriesSampler} to sample a task by sampling one image per thread. In this way, data loading will not be a bottleneck anymore, even though under the condition of a large number of batch (task) size, \textit{i.e.,} multi-task episodic training.

\textbf{Data augmentation} is a nontrivial technique to boost the classification performance in FSL. Most of the existing FSL methods adopt the same data augmentation for both support set and query set at the training stage, but do not apply data augmentation at the test stage. However, recently, some works~\cite{TianWKTI20_RFS,rajasegaran2020self_SKD} have introduced data augmentation into the test stage for the support set, which shows the effectiveness of such an operation. Therefore, both kinds of data augmentation strategies are supported in LibFewShot (\textit{i.e.,} \textit{collate function}).

Note that, \textit{collate function} in LibFewShot can apply the same or different transformations on the support set and query set separately. In addition, LibFewShot also provides some latest data augmentation methods, such as AutoAugment~\cite{cubuk2019autoaugment}, Cutout~\cite{devries2017improved} and RandomAugment~\cite{cubuk2019randaugment}, which are not officially provided by PyTorch but are still useful, and allows users to conveniently define their own data transformation list.

\begin{table*}[!tp]\small 
	\centering
	\caption{Supported functions of LibFewShot compared with other official FSL codes. Methods using for-loops multi-episodes are marked with \dag, which can not use the characteristic of GPU paralleling and will lead to slower computation.}
\begin{tabular}{c|cc|cc|cc|cc}
    \toprule
    & \multicolumn{2}{c}{\textbf{Multi-episodes}} & \multicolumn{2}{c}{\textbf{Multi-GPUs}} & \multicolumn{2}{c}{\textbf{Different-ways \& shots}} & \multicolumn{2}{c}{\textbf{Different Data Augmentations}} \\
    \midrule
    Method & Official & LibFewShot & Official & LibFewShot & Official & LibFewShot & Official & LibFewShot \\
    \midrule
    \textbf{Baseline} & &    &  & \checkmark   &  &  &  & \checkmark   \\
    \textbf{Baseline++} &   &&  &\checkmark  &  &  &  &\checkmark  \\
    \textbf{RFS} &  &   &\checkmark  &\checkmark  &  &  &\checkmark  &\checkmark  \\
    \textbf{SKD} &  &   &\checkmark  &\checkmark  &  &  &\checkmark  &\checkmark  \\
    \textbf{MAML}&\dag  &\dag  &  &\checkmark  &  &  &  &\checkmark  \\
    \textbf{Versa} &\checkmark  &\checkmark  &  &\checkmark  &  &\checkmark  &  &\checkmark  \\
    \textbf{R2D2} &\checkmark  &\checkmark  &   &\checkmark   &\checkmark  &\checkmark  &  &\checkmark  \\
    \textbf{LEO} &  &\checkmark  &  &\checkmark  &  &  &  &\checkmark  \\
    \textbf{MTL} &  &\dag &  &\checkmark  &  &  &  &\checkmark  \\
    \textbf{ANIL} &  &\dag  &  &\checkmark  &  &  &  &\checkmark  \\
    \textbf{ProtoNet} &  &\checkmark  &  &\checkmark  &\checkmark  &\checkmark  &  &\checkmark  \\
    \textbf{RelationNet} &  &\checkmark  &  &\checkmark  &  &\checkmark  &  &\checkmark  \\
    \textbf{CovaMNet} &\dag   &\checkmark  &\checkmark  &\checkmark  &\checkmark  &\checkmark  &  &\checkmark  \\
    \textbf{DN4} &\dag  &\checkmark  &\checkmark  &\checkmark  &  &\checkmark  &  &\checkmark  \\
    \textbf{CAN} &  &\checkmark  &  &\checkmark  &  &\checkmark  &  &\checkmark  \\
\bottomrule
\end{tabular}
\label{tab:functions}
\end{table*}

\subsection{Trainer and Tester}
\textit{Trainer} is the core of LibFewShot, and \textit{Tester} is an enhanced test version of \textit{Trainer}. In the training phase, \textit{Trainer} prepares the training environment by using the configuration information. 
According to the configuration information, \textit{Trainer} initializes the network parameters, creates the optimizer and assigns the GPU and so on. After that, it calls the training, evaluation, and test functions in a loop until the training is completed. The training information, \textit{e.g.} the configuration information, training log, and checkpoints, is also dumped into the disk. In the test phase, \textit{Tester} does similar things, but only calls the test function to calculate the final evaluation criteria.

\subsection{Configs} 
LibFewShot obtains the configuration information from the YAML file, in which the network structure, episode size, data root, and training epochs are determined. In order to avoid missing some important parameters, we set a default configuration file, and the framework will read this file first. In addition, our framework can also support a user-defined configuration file, which will replace the same parameters in the default configuration file. If some parameters are not defined in the users' configuration file, the framework will use the default profile settings for training.

\subsection{How to Run the LibFewShot?}
The whole program can be stated by the \textit{run\_trainer} and \textit{run\_tester} scripts. When users have implemented their own methods and the corresponding configuration files, or just use our re-implemented methods and configuration files, they only need to modify the configuration file's path in the \textit{run\_trainer} and then run it. The \textit{run\_trainer} script will parse the configuration file first and overwrite some options in the default configuration, and then pass the configurations to \textit{Trainer} to start the training stage. When the training stage is finished, the users can modify the checkpoints' path in the \textit{run\_tester} and overwrite some options to run a test. \textit{Tester} will automatically use the configuration file in the checkpoints directory to set up a network. Note that, the parameters at the test stage can also be overwritten by a manually defined parameter list.

\subsection{Other Supported Functions}
Based on the above designs, LibFewShot can already support multiple advanced functions, including \textit{multi-episodes}, \textit{multi-GPUs}, \textit{different-ways $\&$ shots} and \textit{different data augmentations} for all re-implemented FSL methods. Multi-episodes and multi-GPUs mean that LibFewShot supports multi-task episodic training and multi-GPUs training for each method, respectively. Different-ways $\&$ shots indicate that LibFewShot supports different numbers of ways and shots in the training and evaluation phases. Different data augmentations mean that LibFewShot can support using more flexible data augmentation for the support set and query set. An overview of the comparison between LibFewShot and other FSL methods is shown in Table~\ref{tab:functions}.

\begin{table}[!tp]\small
	\centering
	\caption{Average train/interface time for each task and memory for each GPU, respectively, when using DN4 in LibFewShot.}
	\begin{tabular}{lcccc}
	\toprule
	\textbf{GPUs} &   \textbf{1}  &   \textbf{2}  &   \textbf{4}  &   \textbf{8}\\
	\midrule
	\textbf{Train}   &  $16.07\text{ms}$    &  $12.17\text{ms}$   &  $8.48\text{ms}$    &  $6.84\text{ms}$\\
	\textbf{Inference}   &  $12.04\text{ms}$    &  $6.57\text{ms}$   &  $4.27\text{ms}$    &  $3.04\text{ms}$\\
	\textbf{Memory}$_\text{mean}$ &  $15263\text{MB}$     &  $15425\text{MB}$   &  $15430\text{MB}$     &  $15691\text{MB}$\\
	\textbf{Memory}$_\text{min}$  &  $15263\text{MB}$     &  $15207\text{MB}$   &  $15425\text{MB}$     &  $15407\text{MB}$\\
	\textbf{Memory}$_\text{max}$  &  $15263\text{MB}$     &  $15633\text{MB}$   &  $15957\text{MB}$     &  $17355\text{MB}$\\
	\bottomrule
	\end{tabular}
	\label{tab:libfewshot_multiGPU}
\end{table}

\section{Multi-GPUs}
LibFewShot adopts DataParallel provided in PyTorch to provide the multi-GPUs processing ability. Moreover, LibFewShot not only supports the backbones to train in parallel, but also enables the classifiers to be processed in parallel. Notably, all the re-implemented FSL methods in LibFewShot can be parallelized.

In order to measure the efficiency of multi-GPUs training in LibFewShot, we randomly sample images from \textit{mini}ImageNet and use these images to construct $160,000$ $5$-Way $1$-Shot tasks. Specifically, DN4 is selected to process these tasks, and we calculate the training/interface time and memory distributions. For fairness, we use different episode sizes when using different numbers of GPUs to make sure $16$ tasks are in $1$ GPU. The time and memory consumed by different GPUs during training are shown in Table~\ref{tab:libfewshot_multiGPU}.

As seen, when the number of GPUs increases, the average memory occupied by each task will also increase. This is because the communication between GPUs will also increase the time cost during the multi-GPUs training. However, correspondingly, because more GPUs can be used to train more tasks at the same time, the training speed will be significantly improved. When $8$ GPUs are used, and each GPU has $16$ few-shot tasks, the training speed can reach more than $2$ times faster than only using $1$ GPU.

\end{document}